\PassOptionsToPackage{table}{xcolor}
\documentclass[11pt]{article}

\usepackage[final]{acl}

\usepackage{times}
\usepackage{latexsym}

\usepackage[T1]{fontenc}

\usepackage[utf8]{inputenc}

\usepackage{microtype}

\usepackage{inconsolata}

\usepackage{graphicx}
\usepackage{enumitem}

\usepackage[linesnumbered,ruled,vlined]{algorithm2e}
\usepackage[table]{xcolor}       
\usepackage{amsmath}
\usepackage{tabularx}
\usepackage{boldline}
\usepackage{booktabs}
\usepackage{arydshln}
\usepackage{multirow}
\usepackage{colortbl}
\usepackage{svg}
\usepackage{makecell}
\usepackage{listings}
\usepackage[most]{tcolorbox}

\title{Mind the Perspective: Let's Reason Recursively for Theory of Mind}

\author{
Chao Lei$^{1}$, Guang Hu$^{1}$, Meng Yang$^{2}$, 
Yanbei Jiang$^{1}$, Nir Lipovetzky$^{1}$ \\
$^{1}$ School of Computing and Information Systems, The University of Melbourne, Australia \\
$^{2}$ SensiLab, Monash University, Australia \\
\texttt{\{clei1, yanbeij\}@student.unimelb.edu.au} \\
\texttt{Meng.Yang@monash.edu} \\
\texttt{\{guang.hu, nir.lipovetzky\}@unimelb.edu.au}
}

\begin{document}
\maketitle
\begin{abstract}
Theory of Mind (ToM) reasoning requires inferring agents' beliefs from partial and asymmetric observations, which remains an open challenge for LLMs. 
Existing prompting-based approaches improve ToM reasoning through observable-event filtering or temporal belief chains, without explicitly modeling nested beliefs. 
We introduce \textsc{RecToM}, an inference-time framework for ToM reasoning that models nested beliefs via recursive perspective construction. 
\textsc{RecToM} constructs each character perspective from the preceding character perspective along the character chain specified by the question, reducing higher-order belief questions to actual-world questions within the final constructed perspective. We further provide a KD45 analysis showing that \textsc{RecToM}'s perspective construction induces a well-formed belief modality beyond simple event filtering.
Experiments on ToM benchmarks, including Hi-ToM, Big-ToM, and FanToM, across multiple LLM backbones show that \textsc{RecToM} consistently outperforms recent advanced approaches, achieving state-of-the-art performance.
Notably, \textsc{RecToM} reaches 100\% accuracy on Hi-ToM with GPT-5.4 and Qwen3.5, a benchmark requiring higher-order ToM reasoning. Our code is available at \url{https://github.com/you68681/rectom}.
\end{abstract}

\section{Introduction}

Theory of Mind (ToM), the ability to reason about others' beliefs, knowledge, and perspectives, is a central component of social intelligence \citep{Premack_Woodruff_1978,wimmer1983beliefs,baron1985does}. 
For Large Language Models (LLMs) in interactive settings, ToM is central to handling asymmetric information, multi-agent coordination, and belief-dependent decision making \citep{rabinowitz2018machine,sap2022neural,gandhi2023understanding}. 
However, recent studies show that even strong LLMs remain unreliable on ToM tasks that require reconstructing agent-specific beliefs from partial observations, rather than predicting the final world state \citep{sap2022neural,gandhi2023understanding,he2023hitom,kim2023fantom}. 
For instance, in a Sally-Anne-style false-belief paradigm, if Alice leaves after placing an object in a box and Bob later moves it to a drawer, Alice would believe the object remains in the box, whereas predicting the drawer reflects an omniscient-state bias~\citep{wimmer1983beliefs,baron1985does}. Such cases illustrate the epistemic nature of ToM reasoning, where successful reasoning must identify agent-specific observability, preserve beliefs across unobserved intervals, and revise beliefs only under observed relevant evidence.  Higher-order questions, such as the second-order question ``Where does Alice think Bob will search?'', further challenge ToM reasoning since they require nested belief construction, in which Bob's belief must be represented within Alice's perspective rather than inferred as Bob's actual belief \citep{perner1985john,he2023hitom}.

Recent prompting-based methods address parts of this challenge through structured intermediate reasoning. 
\textsc{SimToM} uses a two-stage prompting procedure: it first filters the story to the events observable to each character in question, and then prompts the model to answer the ToM question using only the filtered context \citep{wilf2024think}. 
\textsc{TimeToM} \citep{hou2024timetom} introduces a temporal space by assigning time points to story sentences or dialogue utterances, and constructs a Temporal Belief State Chain (TBSC) for each character. 
It further separates TBSC into self-world beliefs, which record a character's belief, and social-world beliefs, which record beliefs about other characters' actions that may create belief gaps. The former is used for first-order ToM questions, whereas the latter supports higher-order ToM reasoning. 
For higher-order questions, its Time-Aware Belief Solver identifies each character's accessible time points, intersects them as belief-communication periods, and reduces higher-order questions to first-order questions within these periods. \textsc{SimToM} and \textsc{TimeToM} show that ToM reasoning benefits from intermediate representations that specify observable events for each character and the temporal evolution of character beliefs.
However, neither \textsc{SimToM} nor \textsc{TimeToM} explicitly constructs nested beliefs for reasoning over one character's belief within another character's perspective in higher-order ToM questions. See Appendix \ref{Related_Works} for detailed  related work. 

To address this limitation, we introduce \textsc{RecToM}, an inference-time framework that formulates ToM reasoning as recursive symbolic perspective construction for nested belief modeling.
\textsc{RecToM} explicitly models belief states under partial observability, enabling beliefs to persist across unobserved intervals, update under observed evidence, and nest across character perspectives. 
For each ToM task, each narrative statement or dialogue utterance is abstracted into a fact-based symbolic event and classified as either persistent or transient. 
Persistent events introduce or revise ontic facts, such as object locations, character locations, and character presence, whereas transient events, such as communications, claims, and questions, allow belief updates under partial observability. 
\textsc{RecToM} constructs a global state-event sequence by accumulating ontic facts over the event sequence to build fact-based states and pairing each state with its corresponding persistent or transient event.

From the global state-event sequence, \textsc{RecToM} constructs perspectives recursively. 
The global perspective preserves the complete state-event sequence and serves as the initial source perspective. 
For each character specified in the belief question, \textsc{RecToM} constructs the character's perspective in order by completing the character's partial observation over the current source perspective. 
Observable states and events are retained locally, unobservable events are removed, and unobservable states are completed by inheriting the preceding belief state and revising it with the paired observable events. The newly constructed perspective then becomes the source perspective for the next character. 
In this way, nested beliefs are evaluated relative to preceding perspectives rather than the omniscient narrative, reducing higher-order belief questions to actual-world questions within the final constructed perspective, whose final state determines the answer under a closed-world assumption \citep{reiter1981closed}.

When evaluated on well-established ToM benchmarks, including Hi-ToM \citep{he2023hitom}, Big-ToM \citep{gandhi2023understanding}, and FanToM \citep{kim2023fantom}, \textsc{RecToM} consistently outperforms \textsc{SimToM} and \textsc{TimeToM} across multiple LLMs, demonstrating state-of-the-art performance. 
Its advantage is most evident on Hi-ToM, which includes up to fourth-order questions, where \textsc{RecToM} achieves 100\% accuracy on most evaluated LLMs.
These results indicate that recursive perspective construction with explicit fact-based state representations yields robust gains in higher-order and information-asymmetric ToM reasoning. 
We outline our contributions as follows:
\begin{itemize}[noitemsep, topsep=0pt, leftmargin=*]
\item We introduce \textsc{RecToM}, an inference-time framework that formulates ToM reasoning as recursive perspective construction, modeling nested beliefs with respect to preceding character perspectives.
\item We provide a formal analysis showing that \textsc{RecToM}'s perspective construction induces a well-formed belief modality satisfying KD45.

\item We conduct extensive experiments on Hi-ToM, Big-ToM, and FanToM across diverse LLM backbones, showing that \textsc{RecToM} consistently outperforms current state-of-the-art approaches.
\end{itemize}

\section{Problem Formulation}

We define a ToM instance as
\[
\mathcal{I}=(E,q,y), 
\qquad 
E=(e_1,\ldots,e_T),
\]
where $E$ is an ordered event sequence, each event $e_t\in E$ corresponds to a narrative statement or dialogue utterance, $q$ is the question, and $y$ is the ground-truth answer. 

\begin{figure}[t]
    \centering
    \includegraphics[width=1\linewidth]{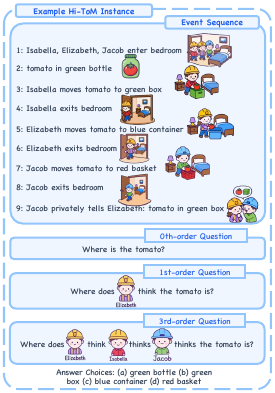}
\caption{An example Hi-ToM instance with zero-order, first-order, and third-order belief questions over a narrative event sequence.}
    \label{fig:hi-tom-example}
\end{figure}

Let $\mathcal{A}(E)$ denote the characters appearing in $E$. 
For a question $q$, we define its character chain as
\[
C(q)=(a_1,\ldots,a_K), \qquad a_i\in\mathcal{A}(E),
\]
where $C(q)$ lists the characters in $q$ from the outermost belief holder $a_1$ to the innermost belief holder $a_K$, and $K$ denotes the belief-nesting order. Figure~\ref{fig:hi-tom-example} illustrates a Hi-ToM instance, where each $e_t$ is a narrative statement. The zero-order (actual-world) question has $K=0$ and $C(q)=\emptyset$; the first-order belief question has $K=1$ and $C(q)=(\mathrm{Elizabeth})$; and the third-order belief question, a higher-order case with $K>1$, has $K=3$ with $C(q)=(\mathrm{Elizabeth}, \mathrm{Isabella}, \mathrm{Jacob})$. 

\definecolor{StepOne}{HTML}{FDBA00}
\definecolor{StepTwo}{HTML}{334155}
\definecolor{StepThree}{HTML}{2563EB}

\begin{figure*}[!t]
    \centering
    \includegraphics[width=1\linewidth]{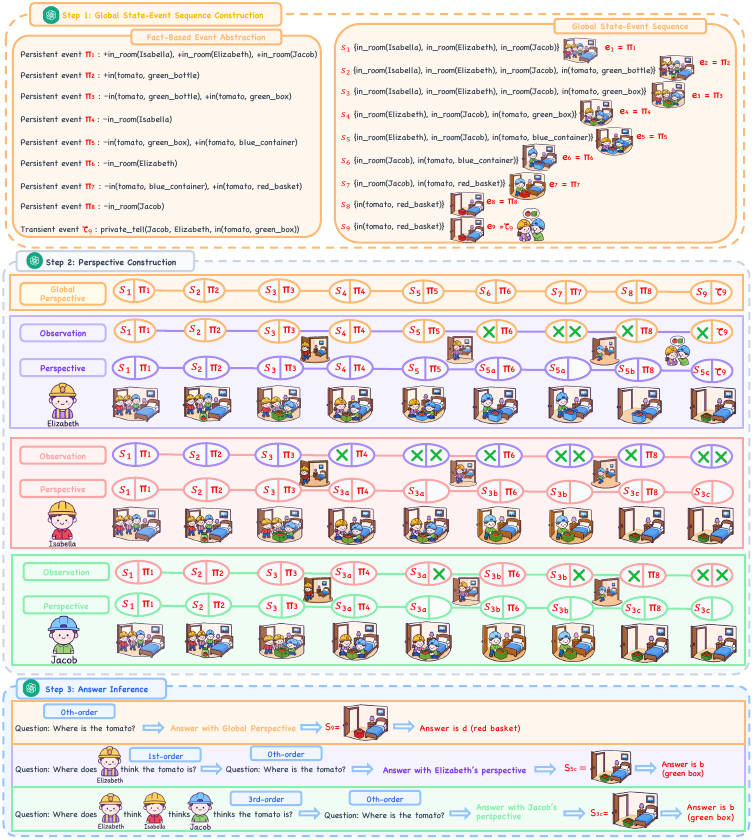}

\caption{Illustration of the full \textsc{RecToM} procedure for solving the Hi-ToM instance in Figure~\ref{fig:hi-tom-example}. Narrative events are abstracted into fact-based persistent events $\pi_i$ and transient events $\tau_i$, and accumulated into a global state-event sequence $\{(s_t,e_t)\}_{t=1}^{T}$, with $T=9$ in this example (\textbf{\textcolor{StepOne}{Step~1}}). Character perspectives are recursively constructed along the character chain $(\mathrm{Elizabeth},\mathrm{Isabella},\mathrm{Jacob})$ (\textbf{\textcolor{StepTwo}{Step~2}}). Crosses mark unobservable states or events, while changes in the suffixes of state labels, such as from $s_{5a}$ to $s_{5b}$, denote belief-state updates induced by observable events during partial-observation completion. Each constructed perspective serves as the source for constructing the perspective of the next character. The zero-order, first-order, and third-order questions are answered from the final states of the global, Elizabeth's, and Jacob's perspectives, respectively, after reducing belief questions to zero-order questions within their corresponding perspectives (\textbf{\textcolor{StepThree}{Step~3}}).}

    \label{fig:RECTOM}
\end{figure*}

\section{\textsc{RecToM} Overview}

\textsc{RecToM} consists of three main steps: 1) global state-event sequence construction; 
2) recursive perspective generation for the characters in $C(q)$; and 3) answer inference from the constructed perspective. The overall procedure of \textsc{RecToM} is illustrated in Figure \ref{fig:RECTOM}.

\subsection{Global State-Event Sequence Construction}

\subsubsection{{Fact-Based Event Abstraction.}}
Given the event sequence $E$, \textsc{RecToM} first abstracts each event $e_t \in E$ into a fact-based symbolic representation and classifies it as either a persistent event $\pi_t$ or a transient event $\tau_t$. A persistent event $\pi_t$ introduces or revises ontic facts about world conditions, such as object locations, character locations, and character presence, which remain valid until explicitly revised by later persistent events. In contrast, a transient event $\tau_t$ records an event occurrence, such as a communication, claim, or question, that updates characters' beliefs by modifying the facts when they are  unobservable. We note that \textsc{RecToM} generates the fact-based representation using the structured abstraction prompt and determines $\pi_t$ and $\tau_t$ according to the task description. For concision, we use $e_t$ to denote the abstracted fact-based event in the following sections, unless explicitly noted.

\subsubsection{Global State-Event Sequence}
\textsc{RecToM} constructs the symbolic state $s_t$ for each event $e_t$ by updating $s_{t-1}$ with added facts $\Delta_t^{+}$ and removed facts $\Delta_t^{-}$, initialized with $s_0=\emptyset$: \begin{equation} s_t=(s_{t-1}\!\setminus\!\Delta_t^-)\cup\Delta_t^+, \; t=1,\ldots,T. \label{eq:state-update} \end{equation} When $e_t=\pi_t$, $\Delta_t^{+}$ and $\Delta_t^{-}$ are specified by the persistent event; when $e_t=\tau_t$, $\Delta_t^{+}=\Delta_t^{-}=\emptyset$, yielding $s_t=s_{t-1}$.
Thus, $s_t$ accumulates the ontic facts that hold after processing $e_t$, providing a complete state representation for belief modeling. \textsc{RecToM} preserves each abstract event $e_t$ at its original position, resulting in the global state-event sequence:
\[
G=\{(s_t,e_t)\}_{t=1}^{T},
\]
where each pair $(s_t,e_t)$ aligns the symbolic state with its corresponding event, instantiated as either $\pi_t$ or $\tau_t$. \textcolor{StepOne}{Step~1} in Figure~\ref{fig:RECTOM} illustrates the construction of $G$ for the Hi-ToM example in Figure~\ref{fig:hi-tom-example}.

\subsection{Perspective Construction}

\subsubsection{Global Perspective}

\textsc{RecToM} treats the global state-event sequence $G$ as the global perspective, denoted by $P^{(*\mid G)}$. Here, $*$ indicates that the perspective $P$ is not associated with a specific character, and $G$ serves as its source. We note that $G$ contains complete state-event information independent of any character's beliefs or observations. This allows \textsc{RecToM} to answer zero-order questions about real-world conditions, which are encoded in the final state of $P^{(*|G)}$, without character-specific belief inference.

\subsubsection{Character Perspective}
A character perspective encodes the character's belief across the states, combining what the character has observed with what the character continues to believe for unobserved states. 
Given a source perspective $P^{(\cdot)}=\{(s_t,e_t)\}_{t=1}^{T}$, \textsc{RecToM} constructs the perspective of character $a_i$, denoted by $P^{(a_i|P^{(\cdot)})}$, in two steps.

\paragraph{Character Partial Observation.} 
To model what character $a_i$ has observed, \textsc{RecToM} derives a partial observation sequence 
$O^{(a_i|P^{(\cdot)})}=\{(\bar{s}_t,\bar{e}_t)\}_{t=1}^{T}$ over $P^{(\cdot)}$ for $a_i$, where $\bar{s}_t=s_t$ if $s_t$ is observable to $a_i$ and $\bar{s}_t=\emptyset$ otherwise, and $\bar{e}_t=e_t$ if $e_t$ is observable to $a_i$ and $\bar{e}_t=\emptyset$ otherwise. 
Thus, observable states and events in $P^{(\cdot)}$ are retained, unobservable events are discarded, and unobservable states are masked for later completion. \textsc{RecToM} evaluates each pair $(s_t,e_t)$ in $P^{(\cdot)}$ according to the task-specified observability rules.  For example, in Hi-ToM, a state $s_t$ is observable to $a_i$ when it contains presence facts for $a_i$, such as \texttt{in\_room}$(a_i)$; a persistent event $e_t=\pi_t$ is observable if its paired state $s_t$ is observable, while room-entry and room-exit events, such as \texttt{+in\_room(character)} or \texttt{-in\_room(character)}, are observable to all characters; and a transient event $e_t=\tau_t$ is observable when $a_i$ is a participant in the event, such as being the speaker or listener in \texttt{private\_tell(speaker,listener,fact)}.

\paragraph{Partial Observation Completion.} 
To model what character $a_i$ continues to believe for unobserved states, \textsc{RecToM} completes the partial observation sequence of $a_i$.  
In detail, \textsc{RecToM} maps the partial observation sequence $O^{(a_i|P^{(\cdot)})}=\{(\bar{s}_t,\bar{e}_t)\}_{t=1}^{T}$ into the perspective of $a_i$:
\[
P^{(a_i|P^{(\cdot)})}=\{(\hat{s}_t,\bar{e}_t)\}_{t=1}^{T},
\]
where the event component $\bar{e}_t$ is preserved and the state component $\bar{s}_t$ is completed as $\hat{s}_t$: 
\begin{equation}
\hat{s}_t =
\begin{cases}
\bar{s}_t, & \bar{s}_t\neq\emptyset,\\
f_{\mathrm{com}}(\hat{s}_{t-1},\bar{e}_t), & \bar{s}_t=\emptyset.
\end{cases}
\label{eq:perspective-completion}
\end{equation}
Here, the completion function $f_{\mathrm{com}}$ is applied when state is unobservable, i.e., $\bar{s}_t=\emptyset$. It inherits the preceding completed state as the basis of the character's belief to preserve belief continuity, and revises the inherited state with $\bar{e}_t$ when it is observable, i.e., $\bar{e}_t\neq\emptyset$, to account for belief updates. 
This follows the belief-persistence assumption that a character's belief remains unchanged unless revised by observable evidence, consistent with prior work~\cite{hu2023planning,goldman1979justification}. In practice, \textsc{RecToM} updates $\hat{s}_{t-1}$ with $\bar{e}_t$ using Eq.~\eqref{eq:state-update} when $\bar{e}_t=\pi_t$, and revises $\hat{s}_{t-1}$ according to the task-specified transient-event rules when $\bar{e}_t=\tau_t$.

\paragraph{Recursive Perspective Generation} For a belief question of order $K>0$, the global perspective $P^{(*|G)}$ serves as the source perspective for recursively constructing character perspectives along the character chain $C(q)$. 
Each constructed perspective then serves as the source for the next character, thereby modeling nested beliefs relative to the preceding perspective, as shown in \textcolor{StepTwo}{Step~2} of Figure~\ref{fig:RECTOM}.

\subsection{Answer Inference}

Recursive perspective construction reduces first-order and higher-order questions to zero-order questions within the final constructed perspective. 
Under the closed-world assumption, \textsc{RecToM} answers the reduced question using the final state of this perspective, which encodes the innermost character's belief conditioned on the preceding perspectives in $C(q)$, as illustrated in \textcolor{StepThree}{Step~3} of Figure~\ref{fig:RECTOM}.

\SetKwInOut{KwIn}{Input}
\SetKwInOut{KwOut}{Output}

\begin{algorithm}[t]
\small
\caption{Pseudocode of \textsc{RecToM}}
\label{alg:rectom}

\KwIn{Event sequence $E=(e_1,\ldots,e_T)$; ToM question $q$; LLM $\mathcal{M}$}
\KwOut{Predicted answer $\hat{y}$}

\tcc{\textbf{Global Perspective Construction}}

\tcp{Fact-based event abstraction and event-type classification}

$\mathcal{F}=\{(\Delta_t^+,\Delta_t^-, e_t),\ e_t\in\{\pi_t,\tau_t\}\}_{t=1}^{T} \leftarrow \mathcal{M}(E)$\;
\tcp{Construct the global perspective}
$P^{(*|G)} \leftarrow \mathrm{BuildGlobalPerspective}(\mathcal{F})$\;

\tcc{\textbf{Recursive Perspective Construction}}

\tcp{Extract the character chain}
$C(q)=(a_1,\ldots,a_K) \leftarrow \mathcal{M}(q)$\;

\tcp{$C(q)=\emptyset$ indicates a zero-order question}
\If{$C(q)=\emptyset$}{
    \tcp{Answer from the global perspective}
    $\hat{y} \leftarrow \mathcal{M}(q, \mathrm{FinalState}(P^{(*|G)}))$\;
    \Return $\hat{y}$\;
}

\tcp{Initialize source perspective}
$P^{(\cdot)} \leftarrow P^{(*|G)}$\;

\ForEach{$a_i \in C(q)$}{

    \tcp{Build partial observation sequence}
    $O^{(a_i|P^{(\cdot)})} \leftarrow \mathcal{M}(a_i, P^{(\cdot)})$\;

    \tcp{Complete partial observation}
    $P^{(a_i|P^{(\cdot)})} \leftarrow \mathrm{Complete}(O^{(a_i|P^{(\cdot)})}, \mathcal{M})$\;

    \tcp{Update source perspective}
    $P^{(\cdot)} \leftarrow P^{(a_i|P^{(\cdot)})}$\;
}

\tcc{\textbf{Answer Inference}}

\tcp{Reduce to a zero-order question}
$q^{(0)} \leftarrow \mathcal{M}(q, C(q))$\;

\tcp{Answer from the final perspective}
$\hat{y} \leftarrow \mathcal{M}(q^{(0)}, \mathrm{FinalState}(P^{(\cdot)}))$\;

\Return $\hat{y}$\;
\label{algorithm}
\end{algorithm}

\subsection{Pseudocode of \textsc{RecToM}}

Algorithm~\ref{alg:rectom} summarizes the full procedure of \textsc{RecToM}. 
The LLM $\mathcal{M}$ is used for natural-language interpretation and semantic reasoning, including event abstraction and event-type classification (Line~1), character-chain extraction (Line~3), observability judgment (Line~9), transient-event belief revision within $\mathrm{Complete}(\cdot)$ (Line~10), question reduction (Line~12), and answer inference (Lines~5 and 13). 
In contrast, global state accumulation from persistent events during global-perspective construction (Line~2) and persistent-event updates inside $\mathrm{Complete}(\cdot)$ (Line~10) follow the deterministic update rule in Eq.~\eqref{eq:state-update}.
\textsc{RecToM} combines the semantic reasoning capacity of LLMs with deterministic updates to  improve the reliability of belief persistence, belief revision, and recursive perspective construction in ToM reasoning.

\section{KD45 Analysis of \textsc{RecToM}}

We use KD45, a standard modal logic for belief, to show that \textsc{RecToM}'s perspective construction induces a well-formed belief modality beyond simple event filtering~\citep{malcolm1952knowledge,fagin2004reasoning}. 
In this analysis, \(K\) requires a constructed perspective to support standard logical inference, \(D\) requires internal consistency, and \(4\) and \(5\) concern the preservation of beliefs and non-beliefs under repeated perspective construction for the same character. For concision, we provide a proof sketch here and show the full proof in Appendix~\ref{app:kd45-proof}.

Let $\varphi$ and $\psi$ denote belief-query formulas, which may be symbolic facts or nested belief statements. 
For a character $a_i$, let $B_{a_i}\varphi$ denote that $a_i$ believes $\varphi$. 
\textsc{RecToM} interprets $B_{a_i}\varphi$ by constructing $a_i$'s perspective, $P^{(a_i\mid P^{(\cdot)})}$, from a source perspective $P^{(\cdot)}$, and evaluating $\varphi$ inside $P^{(a_i\mid P^{(\cdot)})}$:
\begin{equation}
P^{(\cdot)} \models B_{a_i}\varphi
\quad \text{iff} \quad
P^{(a_i\mid P^{(\cdot)})} \models \varphi.
\label{belief}
\end{equation}
Thus, $\varphi$ is evaluated in the constructed character perspective rather than in the source perspective.

\subsection{Self-Perspective Idempotence}

The key property for proving KD45 is that the perspective is stable under repeated construction for the same character: 
\begin{equation}
P^{(a_i|P^{(a_i|P^{(\cdot)})})}
=
P^{(a_i|P^{(\cdot)})}.
\label{eq:self-perspective-idempotence}
\end{equation}
\textsc{RecToM} satisfies this property because reconstructing the partial observation sequence of $a_i$ from $P^{(a_i|P^{(\cdot)})}$ yields the same partial observation sequence, so the subsequent completion remains the same and produces the same perspective. 

\subsection{KD45 Satisfaction}

For \textsc{RecToM}, \(K\)  holds because the constructed perspective supports standard logical inference (if \(\varphi\rightarrow\psi\) and \(\varphi\) hold then \(\psi\)  holds). According to Eq. \eqref{belief}, \(B_{a_i}(\varphi\rightarrow\psi)\) and \(B_{a_i}\varphi\) holding in  $P^{(\cdot)}$ implies that \(\varphi\rightarrow\psi\) and \(\varphi\) hold in $P^{(a_i|P^{(\cdot)})}$; hence \(\psi\) holds in $P^{(a_i|P^{(\cdot)})}$,  therefore \(B_{a_i}\psi\)  holds in $P^{(\cdot)}$. \(D\) holds because each constructed perspective maintains the consistent symbolic state under the closed-world assumption, so a fact and its negation cannot both hold. \(4\) and \(5\) follow from Eq.~\eqref{eq:self-perspective-idempotence}: reconstructing the same character's perspective returns the same perspective, so both beliefs and non-beliefs are preserved under repeated construction. 
Therefore, the belief modality induced by \textsc{RecToM}'s  perspective construction satisfies KD45.

\definecolor{LightBlue}{RGB}{232,243,255}

\begin{table*}[t]
\centering
\resizebox{\textwidth}{!}{
\setlength{\tabcolsep}{15pt}
\renewcommand{\arraystretch}{1.1}
\begin{tabular}{llcccccc}
\toprule
\textbf{Model}  &\textbf{Method} & \multicolumn{6}{c}{\textbf{Hi-ToM}} \\ \cmidrule(lr){3-8}

 &  & \textbf{0th-Order} & \textbf{1st-Order} & \textbf{2nd-Order} & \textbf{3rd-Order} & \textbf{4th-Order} & \textbf{Overall} \\
\midrule
\multirow{4}{*}{GPT-5.4}
& CoT      & \underline{100.00} & 88.75 & 60.00 & 68.75 & 71.25 & 77.75 \\
& \textsc{SimToM}  & \underline{100.00} & \underline{92.50} & 88.75 & \underline{87.50} & 80.00 & 89.75 \\
& \textsc{TimeToM} & \underline{100.00} & \underline{92.50} & \underline{90.00} & \underline{87.50} & \underline{81.25} & \underline{90.25} \\
\rowcolor{LightBlue}
& \textsc{RecToM} (ours) & \textbf{100.00} & \textbf{100.00} & \textbf{100.00} & \textbf{100.00} & \textbf{100.00} & \textbf{100.00} \\
\midrule
\multirow{4}{*}{Gemini-3}
& CoT      & \underline{100.00} & \underline{100.00} & \underline{92.50} & 85.00 & 83.75 & \underline{92.25} \\
& \textsc{SimToM}  & \underline{100.00} & 91.25 & 90.00 & \underline{88.75} & 86.25 & 91.25 \\
& \textsc{TimeToM} & \underline{100.00} & 78.75 & 86.25 & 86.25 & \underline{87.50} & 87.75 \\
\rowcolor{LightBlue}
& \textsc{RecToM} (ours) & \textbf{100.00} & \textbf{100.00} & \textbf{97.50} & \textbf{97.50} & \textbf{97.50} & \textbf{98.50} \\
\midrule
\multirow{4}{*}{Qwen3.5}
& CoT      & \underline{100.00} & \underline{95.00} & 88.75 & 83.75 & 85.00 & 90.50 \\
& \textsc{SimToM}  & \underline{100.00} & 87.50 & \underline{90.00} & \underline{90.00} & \underline{86.25} & \underline{90.75} \\
& \textsc{TimeToM} & \underline{100.00} & 70.00 & 87.50 & 86.25 & 82.50 & 85.25 \\
\rowcolor{LightBlue}
& \textsc{RecToM} (ours) & \textbf{100.00} & \textbf{100.00} & \textbf{100.00} & \textbf{100.00} & \textbf{100.00} & \textbf{100.00} \\
\midrule
\multirow{4}{*}{Gemma-4}
& CoT      & \underline{100.00} & \underline{93.75} & 65.00 & 62.50 & 58.75 & 76.00 \\
& \textsc{SimToM}  & 98.75 & 91.25 & \underline{81.25} & \underline{73.75} & \underline{73.75} & \underline{83.75} \\
& \textsc{TimeToM} & 97.50 & 70.00 & 60.00 & 63.75 & 60.00 & 70.25 \\
\rowcolor{LightBlue}
& \textsc{RecToM} (ours) & \textbf{100.00} & \textbf{100.00} & \textbf{100.00} & \textbf{100.00} & \textbf{97.50} & \textbf{99.50} \\
\bottomrule
\end{tabular}
}

\caption{Accuracy (\%) on Hi-ToM across question orders and overall performance. Each order contains 80 instances, with 400 tasks in total. The best result for each backbone and metric is in bold. The second-best result is underlined.}
\label{tab:hitom_order_accuracy}
\end{table*}
\section{Experiments}

We first evaluate \textsc{RecToM} on Hi-ToM, a benchmark designed to assess higher-order ToM reasoning over narrative event sequences (Figure \ref{fig:hi-tom-example}). 
We select 400 tasks from Hi-ToM, covering question orders from zero to four, with 80 instances per order. 
Each task requires selecting the correct answer from 15 candidate choices. 
We compare \textsc{RecToM} against current state-of-the-art approaches: Chain-of-Thought prompting (CoT), \textsc{SimToM}, and \textsc{TimeToM}. 
To examine robustness across model families and scales, we evaluate all methods with multiple LLM backbones: proprietary models GPT-5.4~\cite{gpt5.4} and Gemini-3-Flash~\cite{Gemini3}, the open-source dense model Qwen3.5-27B~\cite{qwen35blog}, and the open-source Mixture-of-Experts model Gemma-4-26B-A4B~\cite{Gemma4}. See Appendix \ref{Parameter} for parameter settings.

\definecolor{LightBlue}{RGB}{232,243,255}

\begin{table}[t]
\centering
\resizebox{0.48\textwidth}{!}{
\renewcommand{\arraystretch}{1.1}
\begin{tabular}{llcccc}
\toprule
\textbf{Model}  &\textbf{Method} & \textbf{Big-ToM} & \multicolumn{3}{c}{\textbf{FanToM}} \\
\cmidrule(lr){3-3}\cmidrule(lr){4-6}
 &  & \textbf{Overall} & \textbf{1st-Order} & \textbf{2nd-Order} & \textbf{Overall} \\
\midrule
\multirow{4}{*}{GPT-5.4}
& CoT & \underline{99.00} & 88.95 & 86.71 & 87.96 \\
& \textsc{SimToM} & \underline{99.00} & 91.16 & \underline{95.80} & \underline{93.21} \\
& \textsc{TimeToM} & 97.75 & \underline{92.27} & 92.31 & 92.28 \\
\rowcolor{LightBlue}
& \textsc{RecToM} (ours) & \textbf{99.50} & \textbf{96.13} & \textbf{99.30} & \textbf{97.53} \\
\midrule
\multirow{4}{*}{Gemini-3}
& CoT & \underline{98.75} & 77.35 & \underline{90.21} & 83.02 \\
& \textsc{SimToM} & 98.00 & \underline{82.32} & 89.51 & \underline{85.49} \\
& \textsc{TimeToM} & 92.50 & 74.59 & \textbf{91.61} & 82.10 \\
\rowcolor{LightBlue}
& \textsc{RecToM} (ours) & \textbf{99.00} & \textbf{91.16} & \textbf{91.61} & \textbf{91.36} \\
\midrule
\multirow{4}{*}{Qwen3.5}
& CoT & \underline{96.25} & 78.45 & \underline{85.31} & 81.48 \\
& \textsc{SimToM} & 94.75 & \underline{89.50} & 81.82 & \underline{86.11} \\
& \textsc{TimeToM} & 91.00 & 86.74 & \underline{85.31} & \underline{86.11} \\
\rowcolor{LightBlue}
& \textsc{RecToM} (ours) & \textbf{98.50} & \textbf{92.82} & \textbf{86.01} & \textbf{89.81} \\
\midrule
\multirow{4}{*}{Gemma-4}
& CoT & \underline{93.25} & 63.54 & 75.52 & 68.83 \\
& \textsc{SimToM} & 88.00 & 71.27 & \underline{80.42} & 75.31 \\
& \textsc{TimeToM} & 80.00 & \underline{76.24} & 79.72 & \underline{77.78} \\
\rowcolor{LightBlue}
& \textsc{RecToM} (ours) & \textbf{99.00} & \textbf{87.29} & \textbf{84.62} & \textbf{86.11} \\
\bottomrule
\end{tabular}
}

\caption{Accuracy (\%) on Big-ToM and FanToM. Big-ToM contains 400 instances, while FanToM contains 181 first-order and 143 second-order instances, with 324 instances in total. The best result for each backbone and metric is in bold. The second-best result is underlined.}
\label{tab:bigtom_fantom_order_accuracy}
\end{table}

\subsection{Results on Hi-ToM}

Table~\ref{tab:hitom_order_accuracy} reports Hi-ToM accuracy for zero-order to fourth-order questions, together with overall performance.
\textsc{RecToM} achieves the highest overall accuracy across all LLM backbones, reaching 100.00\% with GPT-5.4 and Qwen3.5, 98.50\% with Gemini-3, and 99.50\% with Gemma-4, demonstrating state-of-the-art performance. Compared with the strongest baseline for each backbone, \textsc{RecToM} yields absolute overall accuracy gains of 9.75\%, 6.25\%, 9.25\%, and 15.75\%, respectively, with the largest improvement over \textsc{SimToM} using Gemma-4.
The order-wise results show that all methods perform strongly on zero-order questions, whereas the baselines generally degrade as the order increases, reflecting the difficulty of inferring nested character beliefs under asymmetric information. 
In contrast, \textsc{RecToM} maintains near-perfect accuracy across question orders, with the lowest order-wise accuracy still reaching 97.50\% on Gemini-3 and Gemma-4.
The few remaining errors arise from LLM semantic interpretation failures, such as incorrect observability identification. 
These results indicate that recursive perspective construction provides a robust mechanism for higher-order ToM reasoning and demonstrate the model-independent nature of \textsc{RecToM}.

\subsection{Generalization across ToM Scenarios}

We further evaluate \textsc{RecToM} on benchmarks, Big-ToM and FanToM, under the same experimental settings as Table~\ref{tab:hitom_order_accuracy} to examine its generality across different ToM scenarios. 
Similar to Hi-ToM, Big-ToM follows the Sally--Anne  paradigm, while presents stories in more natural language and extends beyond object-location changes. Following prior work~\cite{wilf2024think,hou2024timetom}, we evaluate 400 forward-belief questions from Big-ToM, consisting of 200 false-belief and 200 true-belief first-order questions. 
FanToM evaluates ToM reasoning in interactive dialogue scenarios, where characters enter and leave ongoing conversations, creating asymmetric information access and distinct mental states~\cite{quesque2020theory}. 
We evaluate 324 FanToM belief questions, including 181 first-order and 143 second-order questions. Detailed benchmark descriptions for Big-ToM and FanToM are provided in Appendix~\ref{Benchmark_Details}.

The results are reported in Table~\ref{tab:bigtom_fantom_order_accuracy}. 
\textsc{RecToM} achieves the highest overall accuracy for every backbone on both datasets, reaching up to 99.50\% on Big-ToM and 97.53\% on FanToM, with the largest absolute gain of 8.33\% over the second-best method on FanToM with Gemma-4. 
It also obtains the best or tied-best order-specific accuracy on FanToM. These results show that \textsc{RecToM} generalizes from narrative event sequences to interactive dialogue scenarios. For \textsc{RecToM}, errors in FanToM mainly reflect incorrect semantic grounding by LLMs. 
For example, partial exposure, where a character enters the conversation late and hears only the ``tail end'' of a conversation, is incorrectly treated as access to the full preceding dialogue.  Moreover, when later utterances are semantically related to earlier ones, LLMs may incorrectly assume that a character who joined later also knows information mentioned before.

\definecolor{LightBlue}{RGB}{232,243,255}
\definecolor{DarkGreen}{RGB}{0,120,0}

\begin{table}[t]
\centering
\resizebox{0.48\textwidth}{!}{
\renewcommand{\arraystretch}{1.1}
\setlength{\tabcolsep}{3pt}
\renewcommand{\arraystretch}{1.08}
\begin{tabular}{llcccccc}
\toprule
\textbf{Model} & \textbf{Method} 
& \multicolumn{2}{c}{\textbf{Hi-ToM}} 
& \multicolumn{2}{c}{\textbf{Big-ToM}} 
& \multicolumn{2}{c}{\textbf{FanToM}} \\
\cmidrule(lr){3-4}\cmidrule(lr){5-6}\cmidrule(lr){7-8}
& 
& \textbf{Tok} & \textbf{Eff} 
& \textbf{Tok} & \textbf{Eff} 
& \textbf{Tok} & \textbf{Eff} \\
\midrule
\multirow{4}{*}{GPT-5.4}
& CoT 
& 0.6K & -- 
& 0.3K & -- 
& 0.9K & -- \\
& \textsc{SimToM} 
& 2.3K & \textbf{7.3} 
& 0.5K & 0.0 
& 4.0K & \textbf{1.7} \\
& \textsc{TimeToM} 
& 3.0K & 5.4 
& 1.1K & \textcolor{DarkGreen}{-1.5} 
& 5.4K & 1.0 \\
\rowcolor{LightBlue}
& \textsc{RecToM} (ours) 
& 6.6K & 3.8 
& 3.4K & \textbf{0.2} 
& 8.7K & 1.2 \\
\midrule
\multirow{4}{*}{Gemini-3}
& CoT 
& 1.1K & -- 
& 0.4K & -- 
& 1.1K & -- \\
& \textsc{SimToM} 
& 3.7K & \textcolor{DarkGreen}{-0.4} 
& 1.2K & \textcolor{DarkGreen}{-1.0} 
& 5.7K & 0.5 \\
& \textsc{TimeToM} 
& 7.5K & \textcolor{DarkGreen}{-0.7} 
& 3.0K & \textcolor{DarkGreen}{-2.4} 
& 9.4K & \textcolor{DarkGreen}{-0.1} \\
\rowcolor{LightBlue}
& \textsc{RecToM} (ours) 
& 9.4K & \textbf{0.7} 
& 4.8K & \textbf{0.1} 
& 10.2K & \textbf{0.9} \\
\bottomrule
\end{tabular}
}

\caption{Cost analysis in the proprietary backbones GPT-5.4 and Gemini-3. Tok denotes the average token usage per question. K is $10^3$. Eff denotes token efficiency relative to CoT, computed as $\Delta$Acc/$\Delta$Tok, where $\Delta$Acc and $\Delta$Tok are the overall accuracy gain and additional token usage over CoT, respectively. Higher values indicate better token efficiency. Negative values indicate that a method consumes more tokens than CoT while achieving lower overall accuracy.}
\label{tab:token_efficiency}
\end{table}
\subsection{Cost Analysis}

Table~\ref{tab:token_efficiency} compares the average token usage per problem (Tok) and token efficiency (Eff) relative to CoT across the evaluated approaches using GPT-5.4 and Gemini-3 as backbones. 
\textsc{RecToM} shows higher token usage.
However, when evaluated by token efficiency, measured as the overall accuracy gain obtained from each additional 1K tokens relative to CoT, \textsc{RecToM} achieves the highest efficiency across all Gemini-3 settings and on Big-ToM with GPT-5.4. 
It is lower than \textsc{SimToM} on FanToM with GPT-5.4, while efficiency remains comparable (1.7 vs. 1.2). 
\textsc{RecToM} is less token-efficient on Hi-ToM with GPT-5.4, where higher-order questions require deeper perspective construction. 
However, this additional computation supports near-perfect accuracy across all LLM backbones on this challenging benchmark, as reported in Table~\ref{tab:hitom_order_accuracy}.

\definecolor{LightBlue}{RGB}{232,243,255}

\begin{table}[t]
\centering
\resizebox{0.48\textwidth}{!}{
\renewcommand{\arraystretch}{1.1}
\begin{tabular}{llcccccc}
\toprule
\textbf{Backbone} & \textbf{Method} 
& \multicolumn{2}{c}{\textbf{Hi-ToM}} 
& \multicolumn{2}{c}{\textbf{Big-ToM}} 
& \multicolumn{2}{c}{\textbf{FanToM}} \\
\cmidrule(lr){3-4}\cmidrule(lr){5-6}\cmidrule(lr){7-8}
& & \textbf{Acc} & \textbf{$\Delta$} 
  & \textbf{Acc} & \textbf{$\Delta$} 
  & \textbf{Acc} & \textbf{$\Delta$} \\
\midrule

\rowcolor{LightBlue}
\multirow{3}{*}{GPT-5.4}
& \textsc{RecToM} & 100.00 & --  & 99.50 & --  & 97.53 & --  \\
& \textit{w/o}-det & 100.00 & 0.00 & 99.50 & 0.00 & 97.53 & 0.00 \\
& \textit{w/o}-state & 93.00 & 7.00 & 99.25 & 0.25 & 94.44 & 3.09 \\
\midrule

\rowcolor{LightBlue}
\multirow{3}{*}{Gemini-3}
& \textsc{RecToM} & 98.50 & --  & 99.00 & --  & 91.36 & --  \\
& \textit{w/o}-det & 98.00 & 0.50 & 99.00 & 0.00 & 90.43 & 0.93 \\
& \textit{w/o}-state & 94.25 & 4.25 & 98.75 & 0.25 & 87.65 & 3.70 \\
\midrule

\rowcolor{LightBlue}
\multirow{3}{*}{Qwen3.5}
& \textsc{RecToM} & 100.00 & --  & 98.50 & --  & 89.81 & --  \\
& \textit{w/o}-det & 100.00 & 0.00 & 98.50 & 0.00 & 89.50 & 0.31 \\
& \textit{w/o}-state & 93.50 & 6.50 & 97.25 & 1.25 & 87.35 & 2.47 \\
\midrule

\rowcolor{LightBlue}
\multirow{3}{*}{Gemma-4}
& \textsc{RecToM} & 99.50 & --  & 99.00 & --  & 86.11 & --  \\
& \textit{w/o}-det & 98.00 & 1.50 & 98.75 & 0.25 & 85.19 & 0.93 \\
& \textit{w/o}-state & 88.25 & 11.25 & 95.25 & 3.75 & 80.86 & 5.25 \\

\bottomrule
\end{tabular}
}

\caption{Ablation study of \textsc{RecToM}. \textit{w/o}-det replaces deterministic state updates with LLM-based state updating.  \textit{w/o}-state removes symbolic state construction and maintains cumulative observable event sequences in each perspective. Acc denotes overall accuracy (\%). $\Delta$ denotes the accuracy decrease relative to \textsc{RecToM}.}
\label{tab:rectom_ablation}
\end{table}

\subsection{Ablation Study}

To examine the contribution of deterministic state updates and symbolic state representation, we compare \textsc{RecToM} with two ablated variants. 
The \textit{w/o}-det variant retains symbolic states, while replacing  deterministic state updates with LLM-based execution under the same update rule in Eq.~\eqref{eq:state-update}. 
The \textit{w/o}-state variant removes state representation from each perspective while preserving recursive perspective construction along $C(q)$. 
Instead of constructing perspectives through observation-based completion over state-event pairs, it maintains a cumulative observable event history at each step, $\bar{E}_t=\{\bar{e}_1,\ldots,\bar{e}_t\}$, yielding the perspective $\tilde{P}=\{\bar{E}_t\}_{t=1}^{T}$.

Table~\ref{tab:rectom_ablation} reports the ablation results.  The \textit{w/o}-det variant leads to slight performance degradation, suggesting that deterministic updates reduce variability in belief-state maintenance. Removing symbolic state representation, the \textit{w/o}-state variant, consistently reduces performance across all benchmarks and backbones, with the largest degradation of 11.25\% recorded on Hi-ToM using Gemma-4. These results demonstrate that fact-based state representations provide a more reliable basis for belief reasoning than cumulative observable event histories. 
The \textit{w/o}-state variant requires the LLM to implicitly infer the observability from accumulated event semantics, as well as belief persistence and revision across observed and unobserved events, which can lead to incorrect observability judgments, inconsistent beliefs, and erroneous answer inference.
In contrast, \textsc{RecToM} explicitly constructs fact-based states after each event, encoding belief-relevant conditions through ontic fact updates. The fact-based state representation supports accurate observability evaluation via explicit character presence and location facts and enables answer inference directly from ontic facts in the constructed perspective.  These advantages are evident in higher-order reasoning on Hi-ToM and FanToM, where errors can propagate through recursively constructed perspectives.

\section{Conclusion}
We introduced \textsc{RecToM}, an inference-time framework for ToM reasoning. 
\textsc{RecToM} models nested beliefs by recursively constructing character perspectives, where each constructed perspective becomes the source for constructing the next character's perspective.
This reduces higher-order belief questions to actual-world questions within the perspective of the innermost character specified by the question.
We further provided a KD45 analysis showing that \textsc{RecToM}'s  perspective construction induces a well-formed belief structure beyond simple event filtering. Experiments on Hi-ToM, Big-ToM, and FanToM benchmarks demonstrate that \textsc{RecToM} consistently outperforms recent advanced approaches, across multiple LLM backbones, achieving state-of-the-art performance with the strongest gains on higher-order ToM questions.

\section{Limitations}

\textsc{RecToM} is designed for controlled text-based ToM benchmarks where event-transition rules and observability assumptions are specified by the task description. 
While this setting covers higher-order beliefs, asymmetric information, and dialogue-based belief reasoning, extending \textsc{RecToM} to open-ended or multimodal environments may require additional grounding of implicit observations and event-transition rules.
Moreover, \textsc{RecToM} explicitly constructs character perspectives before deriving the final answer, which introduces additional inference-time computation. 
Future work can reduce this cost through prompt compression, caching shared perspective states, or lightweight symbolic abstraction.

\bibliography{custom}

@article{Premack_Woodruff_1978, 
title={Does the chimpanzee have a theory of mind?}, 
volume={1}, 
number={4}, 
journal={Behavioral and Brain Sciences}, 
author={Premack, David and Woodruff, Guy}, 
year={1978}, 
pages={515–526}}

@inproceedings{sap2019socialiqa,
  title={Social IQa: Commonsense Reasoning about Social Interactions},
  author={Sap, Maarten and Rashkin, Hannah and Chen, Derek and Le Bras, Ronan and Choi, Yejin},
  booktitle={Proceedings of the 2019 Conference on Empirical Methods in Natural Language Processing and the 9th International Joint Conference on Natural Language Processing (EMNLP-IJCNLP)},
  pages={4463--4473},
  year={2019}
}

@misc{ullman2023large,
      title={Large Language Models Fail on Trivial Alterations to Theory-of-Mind Tasks}, 
      author={Tomer Ullman},
      year={2023},
      eprint={2302.08399},
      archivePrefix={arXiv},
      primaryClass={cs.AI},
      url={https://arxiv.org/abs/2302.08399}, 
}

@inproceedings{shapira2024clever,
    title = "Clever Hans or Neural Theory of Mind? Stress Testing Social Reasoning in Large Language Models",
    author = "Shapira, Natalie  and
      Levy, Mosh  and
      Alavi, Seyed Hossein  and
      Zhou, Xuhui  and
      Choi, Yejin  and
      Goldberg, Yoav  and
      Sap, Maarten  and
      Shwartz, Vered",
    booktitle = "Proceedings of the 18th Conference of the European Chapter of the Association for Computational Linguistics (Volume 1: Long Papers)",
    month = mar,
    year = "2024",
    pages = "2257--2273",
}

@inproceedings{kim2023fantom,
    title = "{FANT}o{M}: A Benchmark for Stress-testing Machine Theory of Mind in Interactions",
    author = "Kim, Hyunwoo  and
      Sclar, Melanie  and
      Zhou, Xuhui  and
      Bras, Ronan  and
      Kim, Gunhee  and
      Choi, Yejin  and
      Sap, Maarten",
    booktitle = "Proceedings of the 2023 Conference on Empirical Methods in Natural Language Processing",
    month = dec,
    year = "2023",
    address = "Singapore",
    publisher = "Association for Computational Linguistics",
    pages = "14397--14413"
}

@inproceedings{he2023hitom,
    title = "Hi-{T}o{M}: A Benchmark for Evaluating Higher-Order Theory of Mind Reasoning in Large Language Models",
    author = "Wu, Yufan  and
      He, Yinghui  and
      Jia, Yilin  and
      Mihalcea, Rada  and
      Chen, Yulong  and
      Deng, Naihao",
    booktitle = "Findings of the Association for Computational Linguistics: EMNLP 2023",
    month = dec,
    year = "2023",
    address = "Singapore",
    publisher = "Association for Computational Linguistics",
    pages = "10691--10706",

}

@inproceedings{chen2024tombench,
  title = {ToMBench: Benchmarking Theory of Mind in Large Language Models},
  author = {Chen, Zhuang and Wu, Jincenzi and Zhou, Jinfeng and Wen, Bosi and Bi, Guanqun and Jiang, Gongyao and Cao, Yaru and Hu, Mengting and Lai, Yunghwei and Xiong, Zexuan and Huang, Minlie},
  booktitle = {Proceedings of the 62nd Annual Meeting of the Association for Computational Linguistics (Volume 1: Long Papers)},
  pages = {15959--15983},
  year = {2024},
  address = {Bangkok, Thailand},
  publisher = {Association for Computational Linguistics},
  url = {https://aclanthology.org/2024.acl-long.847/}
}

@inproceedings{xu2024opentom,
    title = "{O}pen{T}o{M}: A Comprehensive Benchmark for Evaluating Theory-of-Mind Reasoning Capabilities of Large Language Models",
    author = "Xu, Hainiu  and
      Zhao, Runcong  and
      Zhu, Lixing  and
      Du, Jinhua  and
      He, Yulan",
    booktitle = "Proceedings of the 62nd Annual Meeting of the Association for Computational Linguistics (Volume 1: Long Papers)",
    month = aug,
    year = "2024",
    address = "Bangkok, Thailand",
    publisher = "Association for Computational Linguistics",
    pages = "8593--8623",
}

@inproceedings{sclar2025explore,
title={Explore Theory of Mind: program-guided adversarial data generation for theory of mind reasoning},
author={Melanie Sclar and Jane Dwivedi-Yu and Maryam Fazel-Zarandi and Yulia Tsvetkov and Yonatan Bisk and Yejin Choi and Asli Celikyilmaz},
booktitle={The Thirteenth International Conference on Learning Representations},
year={2025},
}

@inproceedings{wei2022chain,
author = {Wei, Jason and Wang, Xuezhi and Schuurmans, Dale and Bosma, Maarten and Ichter, Brian and Xia, Fei and Chi, Ed H. and Le, Quoc V. and Zhou, Denny},
title = {Chain-of-thought prompting elicits reasoning in large language models},
year = {2022},
address = {Red Hook, NY, USA},
booktitle = {Proceedings of the 36th International Conference on Neural Information Processing Systems},
articleno = {1800},
numpages = {14},
location = {New Orleans, LA, USA},
series = {NeurIPS }
}

@inproceedings{kojima2022large,
author = {Kojima, Takeshi and Gu, Shixiang Shane and Reid, Machel and Matsuo, Yutaka and Iwasawa, Yusuke},
title = {Large language models are zero-shot reasoners},
year = {2022},
address = {Red Hook, NY, USA},
booktitle = {Proceedings of the 36th International Conference on Neural Information Processing Systems},
articleno = {1613},
numpages = {15},
location = {New Orleans, LA, USA},
series = {NeurIPS}
}

@inproceedings{yao2023tree,
title={Tree of Thoughts: Deliberate Problem Solving with Large Language Models},
author={Shunyu Yao and Dian Yu and Jeffrey Zhao and Izhak Shafran and Thomas L. Griffiths and Yuan Cao and Karthik R Narasimhan},
booktitle={Thirty-seventh Conference on Neural Information Processing Systems},
year={2023},
}

@inproceedings{wilf2024think,
    title = "Think Twice: Perspective-Taking Improves Large Language Models' Theory-of-Mind Capabilities",
    author = "Wilf, Alex  and
      Lee, Sihyun  and
      Liang, Paul Pu  and
      Morency, Louis-Philippe",
    booktitle = "Proceedings of the 62nd Annual Meeting of the Association for Computational Linguistics (Volume 1: Long Papers)",
    month = aug,
    year = "2024",
    address = "Bangkok, Thailand",
    publisher = "Association for Computational Linguistics",
    pages = "8292--8308",
}

@inproceedings{sclar2023minding,
  title = {Minding Language Models' (Lack of) Theory of Mind: A Plug-and-Play Multi-Character Belief Tracker},
  author = {Sclar, Melanie and Kumar, Sachin and West, Peter and Suhr, Alane and Choi, Yejin and Tsvetkov, Yulia},
  booktitle = {Proceedings of the 61st Annual Meeting of the Association for Computational Linguistics (Volume 1: Long Papers)},
  pages = {13960--13980},
  year = {2023},
  address = {Toronto, Canada},
  publisher = {Association for Computational Linguistics},
}

@inproceedings{hou2024timetom,
  title = {TimeToM: Temporal Space is the Key to Unlocking the Door of Large Language Models' Theory-of-Mind},
  author = {Hou, Guiyang and Zhang, Wenqi and Shen, Yongliang and Wu, Linjuan and Lu, Weiming},
  booktitle = {Findings of the Association for Computational Linguistics: ACL 2024},
  pages = {11532--11547},
  year = {2024},
  address = {Bangkok, Thailand},
  publisher = {Association for Computational Linguistics}
}

@inproceedings{jung2024perceptions,
    title = "Perceptions to Beliefs: Exploring Precursory Inferences for Theory of Mind in Large Language Models",
    author = "Jung, Chani  and
      Kim, Dongkwan  and
      Jin, Jiho  and
      Kim, Jiseon  and
      Seonwoo, Yeon  and
      Choi, Yejin  and
      Oh, Alice  and
      Kim, Hyunwoo",
    booktitle = "Proceedings of the 2024 Conference on Empirical Methods in Natural Language Processing",
    month = nov,
    year = "2024",
    address = "Miami, Florida, USA",
    publisher = "Association for Computational Linguistics",
    pages = "19794--19809"
}

@inproceedings{xu2025enigmatom,
    title = "{E}nigma{T}o{M}: Improve {LLM}s' Theory-of-Mind Reasoning Capabilities with Neural Knowledge Base of Entity States",
    author = "Xu, Hainiu  and
      Qi, Siya  and
      Li, Jiazheng  and
      Zhou, Yuxiang  and
      Du, Jinhua  and
      Catmur, Caroline  and
      He, Yulan",
    booktitle = "Findings of the Association for Computational Linguistics: ACL 2025",
    month = jul,
    year = "2025",
    address = "Vienna, Austria",
    publisher = "Association for Computational Linguistics",
    pages = "13598--13622"
}

@inproceedings{baltag1998logic,
author = {Baltag, Alexandru and Moss, Lawrence S. and Solecki, Slawomir},
title = {The logic of public announcements, common knowledge, and private suspicions},
year = {1998},
isbn = {1558605630},
publisher = {Morgan Kaufmann Publishers Inc.},
address = {San Francisco, CA, USA},
booktitle = {Proceedings of the 7th Conference on Theoretical Aspects of Rationality and Knowledge},
pages = {43–56},
numpages = {14},
location = {Evanston, Illinois},
series = {TARK '98}
}

@book{van2007dynamic,
	address = {Dordrecht, Netherland},
	author = {Hans van Ditmarsch and Wiebe van der Hoek and Barteld Kooi},
	editor = {},
	publisher = {Springer},
	title = {Dynamic Epistemic Logic},
	year = {2008}
}

@inproceedings{sileo2023mindgames,
    title = "{M}ind{G}ames: Targeting Theory of Mind in Large Language Models with Dynamic Epistemic Modal Logic",
    author = "Sileo, Damien  and
      Lernould, Antoine",
    booktitle = "Findings of the Association for Computational Linguistics: EMNLP 2023",
    month = dec,
    year = "2023",
    address = "Singapore",
    publisher = "Association for Computational Linguistics",
    pages = "4570--4577",
}

@inproceedings{wu2025deltom,
  title = {DEL-ToM: Inference-Time Scaling for Theory-of-Mind Reasoning via Dynamic Epistemic Logic},
  author = {Wu, Yuheng and Xie, Jianwen and Zhang, Denghui and Xu, Zhaozhuo},
  booktitle = {Proceedings of the 2025 Conference on Empirical Methods in Natural Language Processing},
  pages = {11383--11397},
  year = {2025},
  month = nov,
  address = {Suzhou, China},
  publisher = {Association for Computational Linguistics},
  doi = {10.18653/v1/2025.emnlp-main.573},
  url = {https://aclanthology.org/2025.emnlp-main.573/}
}

@inproceedings{wang2023self,
title={Self-Consistency Improves Chain of Thought Reasoning in Language Models},
author={Xuezhi Wang and Jason Wei and Dale Schuurmans and Quoc V Le and Ed H. Chi and Sharan Narang and Aakanksha Chowdhery and Denny Zhou},
booktitle={The Eleventh International Conference on Learning Representations },
year={2023},
url={https://openreview.net/forum?id=1PL1NIMMrw}
}

@misc{brown2024large,
      title={Large Language Monkeys: Scaling Inference Compute with Repeated Sampling}, 
      author={Bradley Brown and Jordan Juravsky and Ryan Ehrlich and Ronald Clark and Quoc V. Le and Christopher Ré and Azalia Mirhoseini},
      year={2024},
      eprint={2407.21787},
      archivePrefix={arXiv},
      primaryClass={cs.LG},
      url={https://arxiv.org/abs/2407.21787}, 
}

@inproceedings{snell2025scaling,
title={Scaling {LLM} Test-Time Compute Optimally Can be More Effective than Scaling Parameters for Reasoning},
author={Charlie Victor Snell and Jaehoon Lee and Kelvin Xu and Aviral Kumar},
booktitle={The Thirteenth International Conference on Learning Representations},
year={2025},
}

@inproceedings{rabinowitz2018machine,
  title={Machine theory of mind},
  author={Rabinowitz, Neil and Perbet, Frank and Song, Francis and Zhang, Chiyuan and Eslami, SM Ali and Botvinick, Matthew},
  booktitle={International conference on machine learning},
  pages={4218--4227},
  year={2018},
  organization={PMLR}
}

@inproceedings{sap2022neural,
  title={Neural theory-of-mind? on the limits of social intelligence in large lms},
  author={Sap, Maarten and Le Bras, Ronan and Fried, Daniel and Choi, Yejin},
  booktitle={Proceedings of the 2022 conference on empirical methods in natural language processing},
  pages={3762--3780},
  year={2022}
}

@inproceedings{gandhi2023understanding,
author = {Gandhi, Kanishk and Fr\"{a}nken, J.-Philipp and Gerstenberg, Tobias and Goodman, Noah D.},
title = {Understanding social reasoning in language models with language models},
year = {2023},
address = {Red Hook, NY, USA},
booktitle = {Proceedings of the 37th International Conference on Neural Information Processing Systems},
articleno = {595},
numpages = {12},
location = {New Orleans, LA, USA},
series = {NeurIPS}
}

@inproceedings{le2019revisiting,
  title={Revisiting the evaluation of theory of mind through question answering},
  author={Le, Matthew and Boureau, Y-Lan and Nickel, Maximilian},
  booktitle={Proceedings of the 2019 Conference on Empirical Methods in Natural Language Processing and the 9th International Joint Conference on Natural Language Processing (EMNLP-IJCNLP)},
  pages={5872--5877},
  year={2019}
}

@article{perner1985john,
  title={“John thinks that Mary thinks that…” attribution of second-order beliefs by 5-to 10-year-old children},
  author={Perner, Josef and Wimmer, Heinz},
  journal={Journal of experimental child psychology},
  volume={39},
  number={3},
  pages={437--471},
  year={1985},
  publisher={Elsevier}
}

@article{wimmer1983beliefs,
  title={Beliefs about beliefs: Representation and constraining function of wrong beliefs in young children's understanding of deception},
  author={Wimmer, Heinz and Perner, Josef},
  journal={Cognition},
  volume={13},
  number={1},
  pages={103--128},
  year={1983},
  publisher={Elsevier}
}

@article{baron1985does,
  title={Does the autistic child have a “theory of mind”?},
  author={Baron-Cohen, Simon and Leslie, Alan M and Frith, Uta},
  journal={Cognition},
  volume={21},
  number={1},
  pages={37--46},
  year={1985},
  publisher={Elsevier}
}

@inproceedings{hu2023planning,
  title={Planning with multi-agent belief using justified perspectives},
  author={Hu, Guang and Miller, Tim and Lipovetzky, Nir},
  booktitle={Proceedings of the International Conference on Automated Planning and Scheduling},
  volume={33},
  pages={180--188},
  year={2023}
}

@incollection{goldman1979justification,
author="Goldman, Alvin I.",
title="What is Justified Belief?",
bookTitle="Justification and Knowledge: New Studies in Epistemology",
year="1979",
publisher="Springer Netherlands",
address="Dordrecht",
pages="1--23",
isbn="978-94-009-9493-5",
doi="10.1007/978-94-009-9493-5_1",
url="https://doi.org/10.1007/978-94-009-9493-5_1"
}

@article{malcolm1952knowledge,
 URL = {http://www.jstor.org/stable/2252293},
 author = {Norman Malcolm},
 journal = {Mind},
 number = {242},
 pages = {178--189},
 publisher = {[Oxford University Press, Mind Association]},
 title = {Knowledge and Belief},
 urldate = {2026-08-30},
 volume = {61},
 year = {1952}
}

@book{fagin2004reasoning,
  title={Reasoning about knowledge},
  author={Fagin, Ronald and Halpern, Joseph Y and Moses, Yoram and Vardi, Moshe},
  year={1995},
  publisher={The MIT press}
}

@misc{gpt5.4,
  title={Introducing GPT-5.4},
  author={OpenAI},
url={https://openai.com/index/introducing-gpt-5-4/},
  note={Accessed: 2026-05-21},
  year={2026}
}

@misc{Gemini3,
  title={Gemini 3 Developer Guide},
  author={Google DeepMind},
url={https://ai.google.dev/gemini-api/docs/gemini-3},
  note={Accessed: 2026-05-21},
  year={2025}
}

@misc{qwen35blog,
  title  = {Qwen3.5: Towards Native Multimodal Agents},
  author = {{Qwen}},
  year   = {2026},
  url    = {https://qwen.ai/blog?id=qwen3.5},
  note   = {Accessed: 2026-05-21}
}

@misc{Gemma4,
    title = {Gemma 4},
    url = {https://deepmind.google/models/gemma/gemma-4/},
    author = {Google DeepMind},
    note={Accessed: 2026-05-21},
    year = {2026}
}

@article{quesque2020theory,
  title={What do theory-of-mind tasks actually measure? Theory and practice},
  author={Quesque, Fran{\c{c}}ois and Rossetti, Yves},
  journal={Perspectives on psychological science},
  volume={15},
  number={2},
  pages={384--396},
  year={2020},
  publisher={Sage Publications Sage CA: Los Angeles, CA}
}

@inproceedings{zhang2026autotom,
 author = {Zhang, Zhining and Jin, Chuanyang and Jia, Mung Yao and Zhang, Shunchi and Shu, Tianmin},
 booktitle = {Advances in Neural Information Processing Systems},
 editor = {D. Belgrave and C. Zhang and H. Lin and R. Pascanu and P. Koniusz and M. Ghassemi and N. Chen},
 pages = {129659--129699},
 title = {AutoToM: Scaling Model-based Mental Inference via Automated Agent Modeling},
 year = {2025}
}

@incollection{reiter1981closed,
  title={On closed world data bases},
  author={Reiter, Raymond},
  booktitle={Readings in artificial intelligence},
  pages={119--140},
  year={1981},
  publisher={Elsevier}
}
\clearpage

\appendix

\section*{Appendix}

\label{sec:appendix}

\definecolor{LightBlue}{RGB}{232,243,255}

\begin{table*}[t]
\centering
\small
\setlength{\tabcolsep}{5.5pt}
\renewcommand{\arraystretch}{1.6}

\begin{tabular}{
p{0.13\textwidth}
p{0.25\textwidth}
p{0.25\textwidth}
p{0.25\textwidth}
}
\toprule
\textbf{Method}
& \textbf{Intermediate Representation}
& \textbf{Belief-Tracking Mechanism}
& \textbf{Higher-Order / Nested-Belief Mechanism} \\
\midrule

\textsc{SimToM}
& Character-specific filtered context
& Context filtering by character knowledge
& Answering from the filtered context \\

\textsc{SymbolicToM}
& Graphical symbolic belief representations
& Tracking entities' beliefs and beliefs about others
& Tracking beliefs about others' beliefs across higher orders \\

\textsc{TimeToM}
& Temporal Belief State Chains (TBSCs)
& Per-character self-world and social-world belief chains
& Belief-communication periods with a Time-Aware Belief Solver \\

\textsc{PercepToM}
& Character-specific perception information
& Perception-to-belief inference
& No \\

\textsc{EnigmaToM}
& Entity-state knowledge and spatial scene graphs
& Entity-state-based belief tracking across ToM orders
& Iterative perspective masking \\

\textsc{DEL-ToM}
& DEL-grounded belief-update traces
& Step-wise belief-update scoring with a Process Belief Model
& DEL-grounded belief updates for higher-order reasoning \\

\rowcolor{LightBlue}
\textsc{RecToM} (ours)
& Fact-based symbolic state-event sequences
& Belief persistence and updates via partial-observation completion
& Recursive perspective construction from the preceding perspective \\

\bottomrule
\end{tabular}

\caption{
Comparison of structured Theory-of-Mind reasoning methods in terms of intermediate representations, belief-tracking mechanisms, and higher-order or nested-belief mechanisms. ``No'' indicates that the method does not explicitly define the corresponding mechanism.
}
\label{tab:method_comparison}
\end{table*}

\section{Related Work} \label{Related_Works}
Early ToM benchmarks adapted classic false-belief paradigms into text-based question answering. ToMi \citep{le2019revisiting}, for example, evaluates whether models can distinguish reality from an agent’s belief in Sally--Anne-style narratives. SocialIQA~\citep{sap2019socialiqa} broadened the scope from belief tracking to everyday social commonsense, including intents, reactions, and social consequences. However, subsequent evaluations showed that LLMs often lack robust ToM-like reasoning. \citet{sap2022neural} found that large pretrained models underperform humans on social reasoning and false-belief tasks. \citet{ullman2023large} further showed that seemingly minor perturbations to classic ToM scenarios can cause dramatic failures. Moreover, \citet{shapira2024clever} argued that apparent success on ToM-style problems may reflect shallow heuristics rather than stable mental-state reasoning. These findings motivate more controlled, diverse, and leakage-resistant evaluations of ToM in LLMs.

Recent work has therefore developed richer ToM benchmarks that move beyond isolated first-order false-belief questions. Big-ToM \citep{gandhi2023understanding} uses causal templates to generate controlled scenarios involving percepts, beliefs, desires, and actions, enabling systematic tests of forward belief inference, forward action inference, and backward belief inference. FanToM \citep{kim2023fantom} shifts evaluation from passive narratives to multiparty conversations in which characters enter and leave discussions, creating information asymmetries between participants. Its results show that models may answer one question format correctly while failing logically related answerability or information-access questions, revealing inconsistent performance across question types. Hi-TOM \citep{he2023hitom} focuses on higher-order recursive beliefs, extending evaluation to third-order and fourth-order ToM and incorporating public and private deceptive communication. More recent benchmarks further broaden the evaluation landscape: ToMBench \citep{chen2024tombench} introduces a large-scale bilingual benchmark spanning multiple mental-state categories and abilities; OpenToM \citep{xu2024opentom} evaluates longer stories with richer psychological states; and ExploreToM \citep{sclar2025explore} uses program-guided adversarial generation to produce challenging and diverse ToM scenarios. Together, these benchmarks show that robust ToM evaluation requires tracking event access, information asymmetry, and higher-order recursive beliefs.

A parallel line of work studies how to improve LLMs’ ToM reasoning. General-purpose reasoning methods such as chain-of-thought prompting \citep{wei2022chain,kojima2022large}, self-consistency \citep{wang2023self}, and tree-of-thought search \citep{yao2023tree} improve performance on many symbolic and mathematical reasoning tasks, but their effect on ToM is mixed. Several ToM benchmarks report that CoT provides limited gains or can even amplify errors when models adopt an incorrect perspective or propagate a mistaken intermediate belief \citep{gandhi2023understanding,kim2023fantom,he2023hitom}. This suggests that ToM reasoning requires not simply longer rationales, but appropriate intermediate representations for tracking the evolution of agents' beliefs. \textsc{SimToM} \citep{wilf2024think} addresses this by decomposing ToM into perspective-taking and question answering: the model first filters the story to what the target character knows, and then answers from that character’s perspective. This two-stage decomposition substantially improves performance over standard prompting baselines, showing that explicitly constructing character perspectives is beneficial.

Subsequent methods make this intermediate structure more explicit. \textsc{SymbolicToM} \citep{sclar2023minding} introduces a plug-and-play multi-character belief tracker, {using graphical symbolic representations to track entities' beliefs, their estimates of other entities' beliefs, and higher-order beliefs rather than relying on implicit language-model inference alone.} \textsc{TimeToM} \citep{hou2024timetom} further emphasizes temporal structure by constructing a temporal space and per-character Temporal Belief State Chains (TBSCs), distinguishing self-world beliefs from social-world beliefs and using shared belief-communication periods {with a Time-Aware Belief Solver} to support higher-order reasoning. \textsc{PercepToM} \citep{jung2024perceptions} decomposes ToM into perception inference and perception-to-belief inference, showing that LLMs may identify what an agent can perceive while still failing to convert that perception into the correct belief state. \textsc{EnigmaToM} \citep{xu2025enigmatom} extends this direction with {structured entity-state knowledge and spatial scene graphs for belief tracking, together with iterative perspective masking for  perspective-taking in higher-order ToM reasoning}. These studies collectively suggest that ToM failures often arise from insufficient tracking of belief evolution: models struggle to determine which information is observable to each character and whether subsequent events should update that character's beliefs.

Formal and verifier-based approaches provide another route to structured ToM reasoning. Dynamic epistemic logic (DEL) offers a principled formalism for representing belief states, event models, and belief updates in multi-agent settings \citep{baltag1998logic,van2007dynamic}. \textsc{MindGames} \citep{sileo2023mindgames} uses epistemic logic to generate controlled reasoning problems. {\textsc{DEL-ToM}} \citep{wu2025deltom} formalizes ToM as a sequence of {DEL-grounded} belief updates. \textsc{DEL-ToM} trains a Process Belief Model using labels generated by a DEL simulator {to score intermediate belief updates} and applies inference-time scaling to select high-scoring belief traces from multiple LLM-generated candidates. 
This connects ToM reasoning to broader work on process supervision and inference-time scaling, where voting, search, or verifiers are used to select among candidate reasoning traces~\citep{wang2023self,brown2024large,snell2025scaling}. This line of work highlights the importance of intermediate belief-update supervision, yet typically relies on external simulators or verifier-based selection rather than direct inference-time perspective construction.

{Table~\ref{tab:method_comparison} compares representative structured ToM reasoning methods in terms of their intermediate representations, belief-tracking mechanisms, and higher-order or nested-belief mechanisms.}

{\textsc{RecToM} differs from these approaches primarily in how nested beliefs are represented and constructed. \textsc{SimToM} reasons from character-specific filtered contexts; \textsc{SymbolicToM} explicitly tracks entities' beliefs and beliefs about others; \textsc{TimeToM} constructs temporal belief-state chains and uses a Time-Aware Belief Solver to identify belief-communication periods for higher-order reasoning; and \textsc{EnigmaToM} performs iterative perspective masking over structured entity-state information. In contrast, \textsc{RecToM} constructs fact-based symbolic state-event sequences and recursively derives each character perspective from the already constructed perspective of the preceding character in the query chain. Thus, an inner character's belief is evaluated relative to the outer character's constructed perspective rather than directly from the original omniscient narrative. Within each perspective, partial-observation completion preserves belief persistence, with observable events triggering belief updates and unobservable states inheriting the preceding belief state.}

{\textsc{RecToM} is an inference-time neuro-symbolic framework that combines LLM-based semantic interpretation with explicit symbolic belief reasoning. The LLM handles fact-based event abstraction, observability judgments, and semantic reasoning for belief updates, while explicit state-transition and belief-update mechanisms maintain symbolic states and character perspectives. This design separates natural-language interpretation from explicit perspective construction and nested-belief reasoning. Unlike verifier-based approaches such as \textsc{DEL-ToM}, which score and select candidate belief-update traces, \textsc{RecToM} performs answer inference from the recursively constructed perspective itself. Its perspective construction further supports formal analysis of the induced belief modality under KD45.}

\definecolor{LightBlue}{RGB}{232,243,255}

\begin{table}[t]
\centering
\resizebox{0.48\textwidth}{!}{
\begin{tabular}{llccc}
\toprule
\textbf{Model} & \textbf{Method} 
& \multicolumn{3}{c}{\textbf{Big-ToM}} \\
\cmidrule(lr){3-5}
& & \textbf{True-belief} & \textbf{False-belief} & \textbf{Overall} \\
\midrule

\multirow{4}{*}{GPT-5.4}
& CoT & \underline{98.50} & \textbf{99.50} & \underline{99.00} \\
& \textsc{SimToM} & \textbf{99.50} & \underline{98.50} & \underline{99.00} \\
& \textsc{TimeToM} & 96.00 & \textbf{99.50} & 97.75 \\
\rowcolor{LightBlue}
& \textsc{RecToM} & \textbf{99.50} & \textbf{99.50} & \textbf{99.50} \\
\midrule

\multirow{4}{*}{Gemini-3}
& CoT & \underline{98.50} & \textbf{99.00} & \underline{98.75} \\
& \textsc{SimToM} & 98.00 & \underline{98.00} & {98.00} \\
& \textsc{TimeToM} & 92.00 & 93.00 & 92.50 \\
\rowcolor{LightBlue}
& \textsc{RecToM} & \textbf{99.00} & \textbf{99.00} & \textbf{99.00} \\
\midrule

\multirow{4}{*}{Qwen3.5}
& CoT & \underline{93.00} & \textbf{99.50} & \underline{96.25} \\
& \textsc{SimToM} & 92.50 & 97.00 & 94.75 \\
& \textsc{TimeToM} & 84.00 & 98.00 & 91.00 \\
\rowcolor{LightBlue}
& \textsc{RecToM} & \textbf{98.00} & \underline{99.00} & \textbf{98.50} \\
\midrule

\multirow{4}{*}{Gemma-4}
& CoT & \underline{88.00} & \underline{98.50} & \underline{93.25} \\
& \textsc{SimToM} & 78.50 & 97.50 & 88.00 \\
& \textsc{TimeToM} & 69.00 & 91.00 & 80.00 \\
\rowcolor{LightBlue}
& \textsc{RecToM} & \textbf{98.50} & \textbf{99.50} & \textbf{99.00} \\

\bottomrule
\end{tabular}
}
\caption{Accuracy (\%) on Big-ToM forward-belief questions. Big-ToM contains 400 forward-belief instances, with 200 true-belief and 200 false-belief instances. The best result for each backbone and metric is in bold, and the second-best result is underlined.}
\label{tab:bigtom_forward_belief}
\end{table}

\begin{table*}[t]
\centering
\resizebox{\textwidth}{!}{
\begin{tabular}{lllcccccc}
\toprule
\textbf{Model} & \textbf{Method} & \textbf{Effort}
& \textbf{0th-Order}
& \textbf{1st-Order}
& \textbf{2nd-Order}
& \textbf{3rd-Order}
& \textbf{4th-Order}
& \textbf{Overall} \\
\midrule

\multirow{3}{*}{GPT-5.4}
& \multirow{3}{*}{\textsc{RecToM}}
& Low
& 100.00 & 100.00 & 100.00 & 100.00 & 100.00 & 100.00 \\
&
& Medium (default)
& 100.00 & 100.00 & 100.00 & 100.00 & 100.00 & 100.00 \\
&
& High
& 100.00 & 100.00 & 100.00 & 100.00 & 100.00 & 100.00 \\
\midrule

\multirow{3}{*}{Gemini-3}
& \multirow{3}{*}{\textsc{RecToM}}
& Low
& 100.00 & 100.00 & 96.25 & 95.00 & 95.00 & 97.25 \\
&
& Medium
& 100.00 & 100.00 & 97.50 & 97.50 & 96.25 & 98.25 \\
&
& High (default)
& 100.00 & 100.00 & 97.50 & 97.50 & 97.50 & 98.50 \\

\bottomrule
\end{tabular}
}
\caption{Accuracy (\%) of \textsc{RecToM} on Hi-ToM under different reasoning-effort settings. The default reasoning effort is medium for GPT-5.4 and high for Gemini-3.}
\label{tab:hitom_reasoning_effort}
\end{table*}
\definecolor{LightBlue}{RGB}{232,243,255}

\begin{table*}[t]
\centering
\resizebox{\textwidth}{!}{
\setlength{\tabcolsep}{15pt}
\renewcommand{\arraystretch}{1.1}
\begin{tabular}{llcccccc}
\toprule
\textbf{Model} & \textbf{Method} 
& \multicolumn{6}{c}{\textbf{Hi-ToM}} \\ 
\cmidrule(lr){3-8}

& & \textbf{0th-Order} & \textbf{1st-Order} & \textbf{2nd-Order} 
& \textbf{3rd-Order} & \textbf{4th-Order} & \textbf{Overall} \\
\midrule

\multirow{4}{*}{GPT-5.4}
& CoT
& \textbf{100.00} & \underline{88.75} & 60.00 & \underline{68.75} & \underline{71.25} & \underline{77.75} \\
& \textsc{SymbolicToM}
& 93.75 & 82.50 & \underline{67.50} & 56.25 & 48.75 & 69.75 \\
& \textsc{AutoToM}
& 98.75 & 87.50 & 66.25 & 61.25 & 55.00 & 73.75 \\
\rowcolor{LightBlue}
& \textsc{RecToM} (ours)
& \textbf{100.00} & \textbf{100.00} & \textbf{100.00} & \textbf{100.00} & \textbf{100.00} & \textbf{100.00} \\
\midrule

\multirow{4}{*}{Qwen3.5}
& CoT
& \textbf{100.00} & \underline{95.00} & \underline{88.75} & 83.75 & \underline{85.00} & \underline{90.50} \\
& \textsc{SymbolicToM}
& 98.75 & 88.75 & 75.00 & 66.25 & 61.25 & 78.00 \\
& \textsc{AutoToM}
& 98.75 & 88.75 & 86.25 & \underline{85.00} & 81.25 & 88.00 \\
\rowcolor{LightBlue}
& \textsc{RecToM} (ours)
& \textbf{100.00} & \textbf{100.00} & \textbf{100.00} & \textbf{100.00} & \textbf{100.00} & \textbf{100.00} \\

\bottomrule
\end{tabular}
}

\caption{Accuracy (\%) on Hi-ToM with additional training-free baselines, including \textsc{SymbolicToM} and \textsc{AutoToM}. Each order contains 80 instances, with 400 tasks in total. The best result for each backbone and metric is in bold, and the second-best result is underlined.}
\label{tab:hitom_additional_baselines}
\end{table*}

\definecolor{LightBlue}{RGB}{232,243,255}

\begin{table}[t]
\centering
\resizebox{0.48\textwidth}{!}{
\setlength{\tabcolsep}{15pt}
\renewcommand{\arraystretch}{1.1}
\begin{tabular}{llc}
\toprule
\textbf{Model} & \textbf{Method} 
& \textbf{Big-ToM Overall} \\
\midrule

\multirow{3}{*}{GPT-5.4}
& CoT & \underline{99.00} \\
& \textsc{AutoToM} & 98.50 \\
\rowcolor{LightBlue}
& \textsc{RecToM} (ours) & \textbf{99.50} \\
\midrule

\multirow{3}{*}{Qwen3.5}
& CoT & \underline{96.25} \\
& \textsc{AutoToM} & 95.00 \\
\rowcolor{LightBlue}
& \textsc{RecToM} (ours) & \textbf{98.50} \\

\bottomrule
\end{tabular}
}

\caption{Accuracy (\%) on Big-ToM with \textsc{AutoToM} included as an additional training-free baseline. The best result for each backbone is in bold, and the second-best result is underlined.}
\label{tab:bigtom_additional_baselines}
\end{table}

\section{Additional Experimental Results}

{Table~\ref{tab:bigtom_forward_belief} reports separate true-belief and false-belief accuracies on Big-ToM forward-belief questions, complementing the overall accuracy in Table~\ref{tab:bigtom_fantom_order_accuracy}. \textsc{RecToM} performs better across both belief settings, with clear gains on true-belief questions. For Gemma-4, \textsc{RecToM} outperforms CoT, the second-best method, by 10.50\% on true-belief accuracy. These results show that recursive perspective construction supports reliable reasoning across both belief conditions.}

Table~\ref{tab:hitom_reasoning_effort} reports the performance of \textsc{RecToM} on Hi-ToM under low, medium, and high reasoning-effort settings for GPT-5.4 and Gemini-3. With GPT-5.4, \textsc{RecToM} achieves 100\% accuracy across all question orders and reasoning-effort settings. With Gemini-3, accuracy remains at 100\% on 0th-order and 1st-order questions across all settings, while higher-order accuracy decreases only slightly as reasoning effort is reduced. \textsc{RecToM} decomposes nested-belief reasoning by recursively constructing each character's perspective from the source perspective. This design avoids simultaneous reasoning over multiple nested perspectives, thereby reducing the method's sensitivity to lower reasoning-effort settings.

{We further compare \textsc{RecToM} with two additional training-free ToM methods, \textsc{SymbolicToM} \cite{sclar2023minding} and \textsc{AutoToM} \cite{zhang2026autotom}. \textsc{SymbolicToM} constructs explicit symbolic graphical representations for each character, tracking both the character's own beliefs and their estimates of other characters' beliefs, and extends these representations to higher-order beliefs by iteratively processing the story one sentence at a time. \textsc{AutoToM} instead formulates ToM reasoning as automated Bayesian inverse planning: it first proposes a Bayesian ToM model for the given problem, performs inference over the target mental variables, and iteratively refines the model by adding mental variables or additional timesteps when the inferred posterior remains uncertain. As shown in Table~\ref{tab:hitom_additional_baselines}, both baselines underperform CoT overall and degrade with increasing belief order, while \textsc{RecToM} remains near-perfect across all orders. Their performance degradation mainly arises from difficulties in handling communication events in Hi-ToM, leading to incorrect belief updates across character perspectives.  We then compare \textsc{RecToM} with \textsc{AutoToM} on Big-ToM, omitting \textsc{SymbolicToM} due to its incompatible event representations. As shown in Table~\ref{tab:bigtom_additional_baselines}, \textsc{AutoToM} performs slightly below CoT with both GPT-5.4 and Qwen3.5, whereas \textsc{RecToM} achieves the highest overall accuracy.}

\section{KD45 Proof for \textsc{RecToM}}
\label{app:kd45-proof}

We provide the full derivation of the KD45 axioms for the belief modality induced by \textsc{RecToM}'s recursive perspective construction. The proof has three steps. First, we define the language of belief queries. Second, we define how such queries are evaluated by \textsc{RecToM}. Third, we show that this evaluation satisfies the \(K\), \(D\), \(4\), and \(5\) axioms of KD45. 

Let \(\mathcal{W}\) be the set of well-formed perspectives. 
In this proof, we simplify the notation and write \(P\) for an arbitrary perspective, e.g.,  \(P \in \mathcal{W}\).
We assume that \(\mathcal{W}\) is closed under \textsc{RecToM}'s perspective construction: for each character \(a_i\), the constructed perspective \(P^{(a_i\mid P^{(\cdot)})}\) also belongs to \(\mathcal{W}\), where $P^{(\cdot)} \in \mathcal{W}$ is the source perspective. We define \(F_{a_i}(P^{(\cdot)})=P^{(a_i\mid P^{(\cdot)})}\), where \(F_{a_i}\) maps any source perspective \(P^{(\cdot)}\) to the constructed perspective \(P^{(a_i\mid P^{(\cdot)})}\) of character \(a_i\). For clarity, we omit the source-perspective indicator $(\cdot)$ in the following proof. 

Given an atomic symbolic fact \(x\) and a set of characters \(\mathcal{A}\), we define belief-query formulas using the grammar:
\[
\varphi,\psi ::= x \mid \neg \varphi \mid (\varphi \rightarrow \psi) \mid B_{a_i}\varphi, \quad a_i \in \mathcal{A}.
\]
Here, \(B_{a_i}\varphi\) means that character \(a_i\) believes \(\varphi\). An example of a nested belief is \(B_{a_i}B_{a_j}\varphi\), which means that character \(a_i\) believes that character \(a_j\) believes \(\varphi\). This grammar includes only negation, implication, and belief as primitive operators. Other Boolean connectives can be introduced as abbreviations in the usual classical way; for example, \(\varphi \vee \psi\) abbreviates \(\neg\varphi \rightarrow \psi\), and \(\varphi \wedge \psi\) abbreviates \(\neg(\varphi \rightarrow \neg\psi)\). Thus, the grammar remains compact while still allowing standard Boolean combinations of belief queries.

In the following proof, we fix an arbitrary character \(a_i\) and show that \(B_{a_i}\) satisfies the KD45 axioms.
We write \(P \models \varphi\) to mean that formula \(\varphi\) is satisfied, or evaluates to true, under perspective \(P\). For an atomic symbolic fact \(x\), \(P \models x\) iff \(x\) holds in the final state of \(P\); if \(x\) is absent from the final state, it is treated as false under the closed-world assumption. The primitive Boolean connectives are interpreted classically: \(P \models \neg\varphi\) iff \(P \not\models \varphi\), and \(P \models \varphi \rightarrow \psi\) iff either \(P \not\models \varphi\) or \(P \models \psi\). Connectives such as \(\wedge\) and \(\vee\), when used, inherit their meanings through the abbreviations defined above. The belief operator is interpreted by switching from the current perspective to the constructed perspective of the queried character:
\begin{equation}
P \models B_{a_i}\varphi \quad \text{iff} \quad F_{a_i}(P) \models \varphi .
\label{eq:belief-semantics}
\end{equation}
Thus, evaluating \(B_{a_i}\varphi\) in source perspective \(P\) means: first construct \(a_i\)'s perspective, and then evaluate \(\varphi\) inside that perspective. Nested belief queries are handled by repeatedly applying the same rule.

\paragraph{Self-perspective idempotence.}
The key property is that constructing the same character's perspective twice does not change the result:
\begin{equation}
F_{a_i}(F_{a_i}(P)) = F_{a_i}(P).
\label{eq:self-idempotence}
\end{equation}
This follows from the construction rule in Eq.~\eqref{eq:perspective-completion}. Once \(F_{a_i}(P)\) has already been constructed, the event sequence contains only events observable to \(a_i\) or \(\emptyset\), and every unobservable state has already been completed by inheriting the previous belief state and revising it only with observable evidence. Therefore, reapplying the same construction does not remove any additional information or add any new information. It preserves the same observable event sequence and returns the same completed state sequence.

\subsection{KD45 Axiom Satisfaction}

We now verify the KD45 axioms under the belief semantics in Eq.~\eqref{eq:belief-semantics}. The main idea is simple: \(K\) and \(D\) follow from ordinary classical reasoning inside the constructed perspective \(F_{a_i}(P)\), while \(4\) and \(5\) follow from self-perspective idempotence.

\paragraph{K: Distribution.}
The \(K\) axiom states \(B_{a_i}(\varphi \rightarrow \psi) \rightarrow (B_{a_i}\varphi \rightarrow B_{a_i}\psi)\). Intuitively, this axiom says that belief is closed under implication: if a character's constructed perspective supports an implication and also supports its premise, then it should also support the conclusion. To verify this for \textsc{RecToM}, suppose \(P \models B_{a_i}(\varphi \rightarrow \psi)\) and \(P \models B_{a_i}\varphi\). By Eq.~\eqref{eq:belief-semantics}, both formulas are evaluated inside the same constructed perspective \(F_{a_i}(P)\), so \(F_{a_i}(P) \models \varphi \rightarrow \psi\) and \(F_{a_i}(P) \models \varphi\). Since implication is interpreted classically inside this perspective, \(F_{a_i}(P) \models \psi\). Applying Eq.~\eqref{eq:belief-semantics} again gives \(P \models B_{a_i}\psi\). Therefore, \(K\) holds.

\paragraph{D: Consistency.}
The \(D\) axiom states \(B_{a_i}\varphi \rightarrow \neg B_{a_i}\neg\varphi\). Intuitively, this axiom says that a character's constructed perspective should not support both a statement and its negation. To verify this for \textsc{RecToM}, suppose \(P \models B_{a_i}\varphi\). By Eq.~\eqref{eq:belief-semantics}, this means \(F_{a_i}(P) \models \varphi\). Since \(F_{a_i}(P)\) is a well-formed perspective and formulas are evaluated with classical negation, \(F_{a_i}(P) \not\models \neg\varphi\). Hence \(P \not\models B_{a_i}\neg\varphi\), so \(P \models \neg B_{a_i}\neg\varphi\). Therefore, \(D\) holds.

\paragraph{4: Positive introspection.}
The \(4\) axiom states \(B_{a_i}\varphi \rightarrow B_{a_i}B_{a_i}\varphi\). Intuitively, this axiom says that if a character's constructed perspective supports \(\varphi\), then recursively asking what the same character believes should not change that perspective. To verify this for \textsc{RecToM}, suppose \(P \models B_{a_i}\varphi\). Then \(F_{a_i}(P) \models \varphi\). By self-perspective idempotence (Eq.~\eqref{eq:self-idempotence}), applying the same construction again gives \(F_{a_i}(F_{a_i}(P))=F_{a_i}(P)\). Thus, when \(B_{a_i}\varphi\) is evaluated inside \(F_{a_i}(P)\), it returns to the same constructed perspective where \(\varphi\) already holds. Hence \(F_{a_i}(P) \models B_{a_i}\varphi\), and by Eq.~\eqref{eq:belief-semantics}, \(P \models B_{a_i}B_{a_i}\varphi\). Therefore, \(4\) holds.

\paragraph{5: Negative introspection.}
The \(5\) axiom states \(\neg B_{a_i}\varphi \rightarrow B_{a_i}\neg B_{a_i}\varphi\). Intuitively, this axiom says that if \(\varphi\) is not supported by a character's constructed perspective, then recursively querying that same character's belief should not make \(\varphi\) become supported. To verify this for \textsc{RecToM}, suppose \(P \models \neg B_{a_i}\varphi\). Then \(F_{a_i}(P) \not\models \varphi\). By self-perspective idempotence (Eq.~\eqref{eq:self-idempotence}), applying the same construction again still yields the same perspective, so \(F_{a_i}(F_{a_i}(P)) \not\models \varphi\). Therefore, inside \(F_{a_i}(P)\), the character does not believe \(\varphi\), i.e., \(F_{a_i}(P) \models \neg B_{a_i}\varphi\). Applying Eq.~\eqref{eq:belief-semantics} again gives \(P \models B_{a_i}\neg B_{a_i}\varphi\). Therefore, \(5\) holds.

Since the belief operator \(B_{a_i}\) satisfies \(K\), \(D\), \(4\), and \(5\), \textsc{RecToM}'s recursive perspective construction induces a KD45 belief modality. 

\section{Parameter Settings}
\label{Parameter}
\textsc{RecToM} operates on all LLMs through API access. For proprietary models, GPT-5.4 and Gemini-3 Flash, we use the default parameter settings. For open-source models, we set temperature to 1.0 and top-$p$ to 0.95 for both Qwen3.5-27B and Gemma-4-26B-A4B, with top-$k$ set to 20 for Qwen3.5-27B and 64 for Gemma-4-26B-A4B to reduce repetitive outputs and filter rare tokens while preserving generation diversity. All experiments are conducted on a virtual machine with four NVIDIA A100 80GB GPUs.

\section{Benchmark Details} \label{Benchmark_Details}

\subsection{Big-ToM}

\begin{figure}[!t]
    \centering
    \includegraphics[width=1\linewidth]{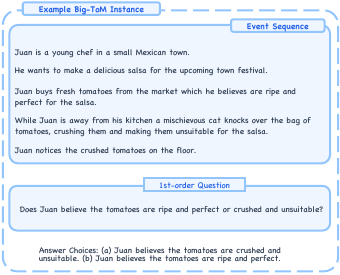}
\caption{An example Big-ToM instance with the first-order belief question over a natural-language event sequence.}
    \label{fig:bigtom-example}
\end{figure}

Big-ToM~\citep{gandhi2023understanding}, a GPT-4-generated benchmark, evaluates belief reasoning in natural-language stories based on the Sally--Anne false-belief paradigm. 
We use the forward-belief subset, which asks models to infer what a character believes after the character either observes or misses a belief-relevant event. This subset focuses on first-order true-belief and false-belief questions, evaluating whether a character's belief is consistent or inconsistent with reality. 
The task is formulated as binary multiple choice, with a random baseline accuracy of 50\%. Figure~\ref{fig:bigtom-example} shows an example true-belief instance from the Big-ToM forward-belief subset.

\subsection{FanToM}

\begin{figure*}[t]
    \centering
    \includegraphics[width=1\linewidth]{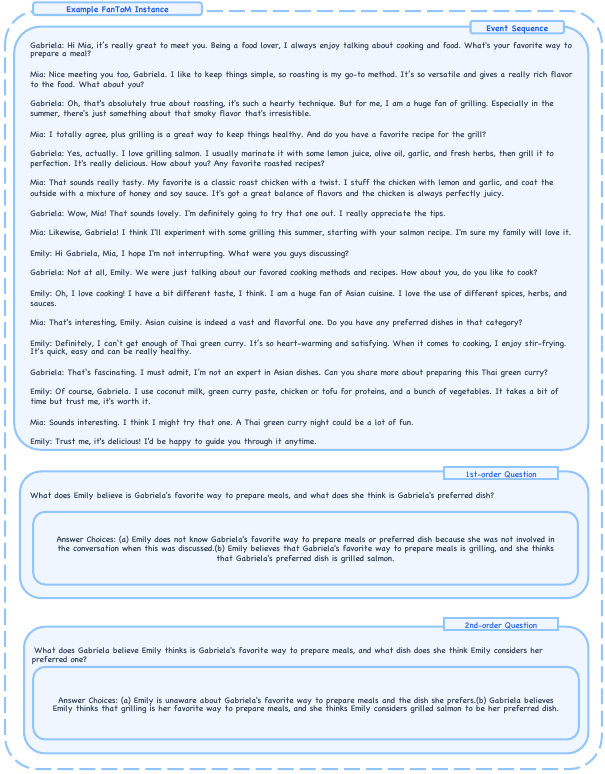}
\caption{An example FanToM instance with first-order and second-order belief questions over a dialogue sequence.}
    \label{fig:fantom-example}
\end{figure*}

FanToM~\citep{kim2023fantom} evaluates ToM reasoning in multi-party dialogue settings. 
Its dialogues introduce asymmetric information by allowing characters to enter and leave while the conversation continues, resulting in different characters observing different parts of the conversation. 
We focus on FanToM belief questions, including first-order and second-order questions, which align with the scope of this work. 
These questions are formulated as binary multiple-choice tasks, with a random baseline accuracy of 50\%. 
Compared with narrative benchmarks, FanToM dialogues are longer and involve more characters and subtopics, requiring models to integrate extended dialogue context and maintain character-specific beliefs. 
Figure~\ref{fig:fantom-example} illustrates the FanToM dialogue structure with first-order and second-order belief questions.

\begin{figure}[t]
    \centering
    \includegraphics[width=1\linewidth]{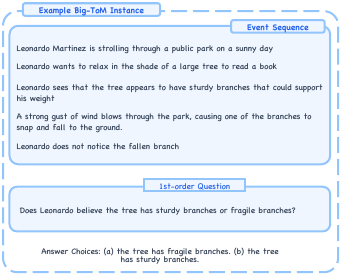}
\caption{An example Big-ToM instance where \textsc{RecToM} makes an incorrect observability judgment.}
    \label{fig:bigtom-incorrect}
\end{figure}

\section{Error Analysis}

The explicit state-event representation and recursive perspective construction in \textsc{RecToM} provide a structured basis for nested-belief reasoning. Nevertheless, semantic misinterpretations by the LLM can still lead to reasoning failures. For example, Figure~\ref{fig:bigtom-incorrect} illustrates a Big-ToM instance in which \textsc{RecToM} produces an incorrect answer due to an erroneous observability judgment by the LLM. Leonardo first observes that a tree appears to have sturdy branches. Later, a branch snaps and falls, while the narrative explicitly states that Leonardo does not notice this event. Leonardo's perspective should therefore preserve the earlier belief that the tree has sturdy branches. However, the LLM treats the branch-falling event as observable to Leonardo, resulting in an erroneous belief update and final answer.

\begin{figure*}[t]
    \centering
    \includegraphics[width=1\linewidth]{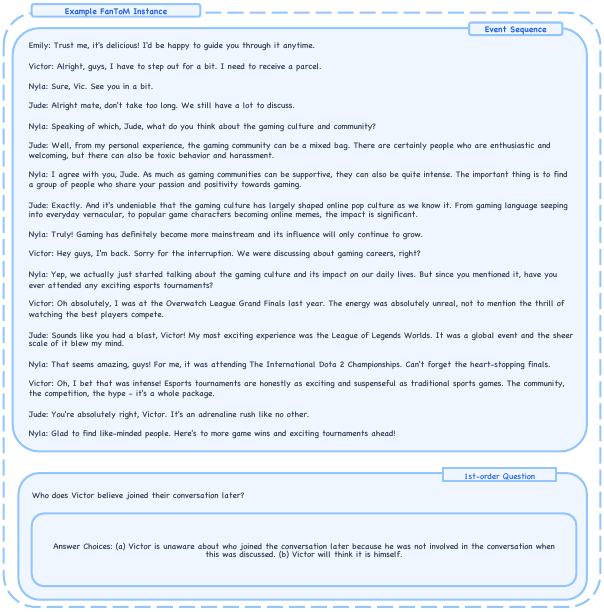}
\caption{An example FanToM instance where \textsc{RecToM} incorrectly grounds a natural-language question to the relevant symbolic facts.}
    \label{fig:fantom-incorrect}
\end{figure*}

{Figure~\ref{fig:fantom-incorrect} illustrates a FanToM example in which \textsc{RecToM} constructs the state-event sequence and character perspective correctly, but the final inference step fails to map the natural-language question to the relevant symbolic fact. Victor leaves the conversation, Nyla and Jude continue talking during his absence, and Victor later returns. The question asks who Victor believes joined the conversation later, and the correct answer is Victor. The constructed state-event sequence accurately records Victor's status using the facts \texttt{depart(Victor)}, \texttt{return(Victor)}, and \texttt{present(Victor)}, and Victor's observation mask correctly excludes the dialogue during his absence and retains the dialogue when he is present.  However, the LLM fails to associate the phrase ``joined the conversation later'' with \texttt{return(Victor)} in the constructed perspective, resulting in an incorrect answer.}

{These examples show that LLM interpretation errors can occur at different stages of \textsc{RecToM}: incorrect observability judgments affect perspective construction, whereas semantic grounding errors can lead to incorrect final inference even when the constructed perspective is correct.}

\section{Prompt Templates for \textsc{RecToM}}
\label{app:prompt_templates}

Tables~\ref{tab:rectom_delta_prompt}--\ref{tab:rectom_answer_prompt} present the main prompt templates used by \textsc{RecToM} on Hi-ToM. 
They cover fact-based event abstraction, state and event observability identification, transient-event belief revision, and final answer inference. 
For readability, we show compact templates; the complete prompts are provided in the released code.

\begin{table*}[t]
\centering
\begin{tcolorbox}[title={\textbf{Prompt for Fact-based Event Abstraction in \textsc{RecToM}}}, fontupper=\small]
You are extracting step-wise symbolic deltas from a ToM story.

\vspace{1mm}
Return valid JSON only.

\vspace{2mm}
\textbf{Example output schema:}
\begin{verbatim}
{
  "characters": ["Avery", "Charlotte"],
  "steps": [
    {
      "step_index": 1,
      "step_text": "Avery entered the living_room.",
      "event_type": "persistent",
      "added_facts": ["in_room(Avery,living_room)"],
      "removed_facts": []
    }
  ]
}
\end{verbatim}

\vspace{1mm}
\textbf{Important definition.}
\begin{itemize}
    \item Do not output a full state for each step. Output only the delta for each step.
    \item \texttt{event\_type} is either \texttt{persistent} or \texttt{transient}. Persistent events introduce or revise state facts; transient events record communication, claims, or questions.
    \item \texttt{added\_facts} are facts that become true because of the current step.
    \item \texttt{removed\_facts} are facts that stop being true because of the current step.
\end{itemize}

\vspace{1mm}
\textbf{Extraction guidelines.}
\begin{itemize}
    \item Include every story step exactly once and include all human characters appearing in the story.
    \item Extract concise symbolic facts for belief-relevant changes, such as character locations, object locations, communication events, and stated claims.
    \item Remove a fact only when the current step makes it false.
    \item Distinguish spoken claims from actual world facts when communication events appear.
    \item Represent private communication as \texttt{private\_tell(speaker,listener,proposition)} and public communication as \texttt{public\_claim(speaker,proposition)}.
    \item If the spoken content is an object-location statement, represent the proposition as \texttt{in(object,container)}.
\end{itemize}

\vspace{1mm}
\textbf{Story steps:} \texttt{\{story\_steps\}}

\vspace{1mm}
\textbf{Assumptions:} \texttt{\{assumptions\}}
\end{tcolorbox}
\caption{Prompt template used by \textsc{RecToM} for fact-based event abstraction. The LLM extracts symbolic deltas and classifies each event as persistent or transient. The complete state sequence is then computed externally using the deterministic update rule.}
\label{tab:rectom_delta_prompt}
\end{table*}
\begin{table*}[t]
\centering
\begin{tcolorbox}[title={\textbf{Prompt for State Observability Identification in \textsc{RecToM}}}, fontupper=\small]
You are deciding whether a target character is present in each aligned source state.

\vspace{1mm}
Return valid JSON only.

\vspace{2mm}
\textbf{Example output schema:}
\begin{verbatim}
{
  "character": "Alice",
  "observation_basis": [
    "at(Alice,kitchen)",
    "at(Alice,kitchen)",
    "not_observable"
  ]
}
\end{verbatim}

\vspace{1mm}
\textbf{Task.}
\begin{itemize}
    \item Return exactly one string for each aligned source state.
    \item The $n$th string corresponds to the $n$th aligned source state.
    \item Each output string should be either a fact indicating the target character's location or \texttt{not\_observable}.
\end{itemize}

\vspace{1mm}
\textbf{Decision rules.}
\begin{itemize}
    \item If the aligned source state contains a fact indicating that the target character is present in a location, return that fact.
    \item Presence may be expressed by facts such as \texttt{at(character,location)} or \texttt{in\_room(character,location)}.
    \item If no fact indicates the target character's presence, return \texttt{not\_observable}.
    \item Do not return bare location names.
\end{itemize}

\vspace{1mm}
\textbf{Target character:} \texttt{\{target\_character\}}

\vspace{1mm}
\textbf{Aligned source states:} \texttt{\{aligned\_source\_states\}}
\end{tcolorbox}
\caption{Prompt template used by \textsc{RecToM} for state observability identification. Given aligned source states, the LLM derives the observation basis for the target character by identifying explicit presence facts or returning \texttt{not\_observable}.}
\label{tab:rectom_state_observability_prompt}
\end{table*}
\begin{table*}[t]
\centering
\begin{tcolorbox}[title={\textbf{Prompt for Event Observability Identification in \textsc{RecToM}}}, fontupper=\small]
You are deciding whether a target character can observe each aligned event fact.

\vspace{1mm}
Return valid JSON only.

\vspace{2mm}
\textbf{Example output schema:}
\begin{verbatim}
{
  "character": "Alice",
  "observation_basis": [
    "-in_room(Isabella,living_room)",
    "private_tell(Alice,Bob,in(ball,bedroom))",
    "public_claim(Bob,in(ball,kitchen))",
    "not_observable"
  ]
}
\end{verbatim}

\vspace{1mm}
\textbf{Task.}
\begin{itemize}
    \item Return exactly one string for each aligned event fact.
    \item The $n$th string corresponds to the $n$th aligned event fact.
    \item Each output string should be either the exact observable event fact or \texttt{not\_observable}.
\end{itemize}

\vspace{1mm}
\textbf{Decision rules.}
\begin{itemize}
    \item If the aligned event is empty, return \texttt{not\_observable}.
    \item For private communication, such as \texttt{private\_tell(speaker,listener,proposition)}, only the speaker and listener can observe the event.
    \item For public communication, such as \texttt{public\_claim(speaker,proposition)}, all characters can observe the event.
    \item For room-entry or room-exit facts, such as \texttt{+in\_room(character,room)} or \texttt{-in\_room(character,room)}, the event is observable to all characters.
    \item Do not rewrite or normalize the event fact; return it exactly as it appears in the aligned event facts.
\end{itemize}

\vspace{1mm}
\textbf{Target character:} \texttt{\{target\_character\}}

\vspace{1mm}
\textbf{Aligned event facts:} \texttt{\{aligned\_event\_facts\}}
\end{tcolorbox}
\caption{Prompt template used by \textsc{RecToM} for event observability identification. Given aligned event facts, the LLM derives the observation basis for the target character by identifying observable communication events and room-transition events, returning either the exact observable facts or \texttt{not\_observable}.}
\label{tab:rectom_event_observability_prompt}
\end{table*}
\begin{table*}[t]
\centering
\begin{tcolorbox}[title={\textbf{Prompt for Transient-Event Belief Revision in \textsc{RecToM}}}, fontupper=\small]
You are applying an observable transient event to a symbolic belief state.

\vspace{1mm}
Return valid JSON only.

\vspace{2mm}
\textbf{Example output schema:}
\begin{verbatim}
{
  "state_before": [
    "in(lettuce,green_drawer)",
    "in_room(Avery,living_room)"
  ],
  "action": [
    "public_claim(Isabella,in(lettuce,green_bathtub))"
  ],
  "action_added_facts": [
    "in(lettuce,green_bathtub)"
  ],
  "action_removed_facts": [
    "in(lettuce,green_drawer)"
  ],
  "state_after": [
    "in(lettuce,green_bathtub)",
    "in_room(Avery,living_room)"
  ]
}
\end{verbatim}

\vspace{1mm}
\textbf{Task.}
\begin{itemize}
    \item Start from the given state.
    \item Apply the observable transient event to this state.
    \item Output the facts added or removed by the event.
    \item Output the updated state after applying the event.
\end{itemize}

\vspace{1mm}
\textbf{Revision rules.}
\begin{itemize}
    \item The transient event may represent communication, claims, questions, or other belief-relevant actions.
    \item If the event contains an object-location fact, use the fact to revise the current  state.
    \item Preserve state facts that are not affected by the event.
\end{itemize}

\vspace{1mm}
\textbf{State:} \texttt{\{state\}}

\vspace{1mm}
\textbf{Observable transient event:} \texttt{\{observable\_transient\_event\}}
\end{tcolorbox}
\caption{Prompt template used by \textsc{RecToM} for transient-event belief revision. Given the current belief state and an observable transient event, the LLM outputs the event-induced added and removed facts and the resulting updated state.}
\label{tab:rectom_transient_revision_prompt}
\end{table*}
\begin{table*}[t]
\centering
\begin{tcolorbox}[title={\textbf{Prompt for Answer Inference in \textsc{RecToM}}}, fontupper=\small]
You are answering a multiple-choice Theory-of-Mind question.

\vspace{1mm}
Return valid JSON only.

\vspace{2mm}
\textbf{Example output schema:}
\begin{verbatim}
{
  "reasoning_summary": "short summary",
  "predicted_answer": "C"
}
\end{verbatim}

\vspace{1mm}
\textbf{Task.}
\begin{itemize}
    \item Use the provided state facts to answer the question.
    \item Select exactly one option from the candidate choices.
    \item The selected answer should correspond to the facts in the given state.
\end{itemize}

\vspace{1mm}
\textbf{Output constraints.}
\begin{itemize}
    \item \texttt{predicted\_answer} must contain only the option letter, such as \texttt{A}, \texttt{B}, or \texttt{C}.
    \item Do not output the choice text in \texttt{predicted\_answer}.
    \item Do not include additional characters or punctuation in \texttt{predicted\_answer}.
\end{itemize}

\vspace{1mm}
\textbf{State:} \texttt{\{state\}}

\vspace{1mm}
\textbf{Question:} \texttt{\{question\}}

\vspace{1mm}
\textbf{Choices:} \texttt{\{choices\}}
\end{tcolorbox}
\caption{Prompt template used by \textsc{RecToM} for answer selection. Given the final state from the constructed perspective, the LLM selects one candidate option and returns the predicted answer as an option letter.}
\label{tab:rectom_answer_prompt}
\end{table*}
\end{document}